\documentclass[preprint,11pt]{elsarticle}
\makeatletter
\def\ps@pprintTitle{%
  \let\@oddhead\@empty
  \let\@evenhead\@empty
  \def\@oddfoot{\hfill\thepage\hfill}
  \let\@evenfoot\@empty
}
\makeatother
\usepackage{geometry}
\usepackage{float}
\usepackage{verbatim} 
\restylefloat{figure}
\floatstyle{plaintop} 
\restylefloat{table}

\usepackage{amsmath,amssymb,amsfonts}
\usepackage{bm}

\usepackage{graphicx}
\usepackage{subcaption}

\usepackage{longtable}
\usepackage{booktabs}
\usepackage{array}
\usepackage{multirow}
\usepackage{siunitx}

\usepackage{algorithm}
\usepackage{algorithmic}

\usepackage{url}   
\usepackage{tikz}
\newcommand{\circlesym}[1]{%
  \tikz[baseline=(char.base)]\node[draw,circle,inner sep=0.0pt](char){#1};}

\usepackage[hidelinks]{hyperref}

\biboptions{authoryear}

\begin{document}
\begin{frontmatter}

\title{Technical Indicator Networks (TINs): An Interpretable Neural Architecture Modernizing Classical Technical Analysis for Adaptive Algorithmic Trading}

\author[label1]{Longfei Lu}
\ead{aGiant@me.com}



\address[label1]{Independent Researcher, Frankfurt am Main, Germany}

\begin{abstract}
Deep neural networks (DNNs) have transformed fields such as computer vision and natural language processing by employing architectures aligned with domain-specific structural patterns. In algorithmic trading, however, there remains a lack of architectures that directly incorporate the logic of traditional technical indicators. This study introduces Technical Indicator Networks (TINs), a structured neural design that reformulates rule-based financial heuristics into trainable and interpretable modules. The architecture preserves the core mathematical definitions of conventional indicators while extending them to multidimensional data and supporting optimization through diverse learning paradigms, including reinforcement learning. Analytical transformations such as averaging, clipping, and ratio computation are expressed as vectorized layer operators, enabling transparent network construction and principled initialization. This formulation retains the clarity and interpretability of classical strategies while allowing adaptive adjustment and data-driven refinement. As a proof of concept, the framework is validated on the Dow Jones Industrial Average constituents using a Moving Average Convergence Divergence (MACD) TIN. Results validate the effectiveness of the proposed framework and demonstrate its potential for enhancing risk-adjusted performance in trading applications. The findings show that TINs establish a generalizable foundation for interpretable, adaptive, and extensible learning systems in structured decision-making domains. In addition to academic contributions, the framework indicates significant commercial potential, providing the basis for upgrading trading platforms with cross-market visibility and enhanced decision-support capabilities.
\end{abstract}

\begin{keyword}
Technical Indicator Networks (TINs)\sep Interpretable Machine Learning\sep Neural Network Topology\sep Adaptive Algorithmic Trading Strategy
\end{keyword}

\end{frontmatter}

\section{Introduction}
\label{introduction}

Technical indicators have long been fundamental tools in financial analysis, offering structured heuristics for detecting trends, reversals, and momentum patterns in historical price data. Constructs such as moving averages, oscillators, and volatility-based filters are widely employed in both discretionary and algorithmic trading, valued for their transparency, ease of implementation, and empirical robustness across diverse market regimes. These characteristics remain particularly relevant in the context of increasingly complex machine learning applications, where interpretability is a critical requirement.

Recent advances in deep learning have motivated the integration of technical indicators into neural architectures with the aim of improving predictive accuracy and strategy performance. Empirical studies have shown that models incorporating such indicators as structured inputs often outperform purely data-driven approaches. For example, studies have improved technical indicators via parameter optimization or by using them as features in deep learning models, and explored deep learning based stock prediction at scale~\cite{Gulmez2023ESWA,Chudziak2023ESWA,Gajamannage2023ESWA,Liu2024ESWA_NEV,Yun2023ESWA_GA,Kehinde2023ESWA_Scientometric}. Fischer and Krauss \cite{fischer2018} employed Long Short-Term Memory (LSTM) networks for sequential financial data and achieved superior directional forecasting compared to traditional time-series models. Similar results have been reported by Bao et al.\ \cite{bao2017}, Nabipour et al.\ \cite{nabipour2020}, and Ni et al.\ \cite{ni2021}, particularly in short-term or high-frequency contexts. Additional research has explored hybrid frameworks combining neural networks with statistical components \cite{alyahya2021}, refined feature selection methodologies (Kim and Kim, 2021), and reinforcement learning approaches that account for transaction costs and market frictions \cite{ding2022}. More recently, Kwon et al.\ \cite{kwon2022} proposed robust time-series validation protocols for financial model evaluation under realistic trading constraints.

Despite these developments, a structural limitation persists: in most existing frameworks, technical indicators are incorporated as fixed, precomputed inputs. Their internal transformations—such as recursive smoothing, normalized differencing, or dynamic clipping—are treated as static preprocessing steps external to the learning architecture. This separation prevents models from internalizing or adapting the analytical logic embedded in indicator computations, thereby constraining both adaptability and interpretability. To the best of my knowledge, no existing work has reformulated these classical indicators into neural architectures that preserve their canonical computational definitions while enabling parameter adaptation through learning algorithms.

This paper introduces the \textbf{Technical Indicator Networks} (TINs) framework, which addresses this limitation by embedding the computational logic of technical indicators directly into neural network architectures. Each indicator is expressed as a set of composable layer operators corresponding to its mathematical definition, with initial parameters derived from canonical formulas. This initialization preserves the original indicator topology, while subsequent training enables parameter adaptation within the same structural constraints. A collection of \textit{Indicator Network} (IN) sharing a common topological form constitutes a \textit{Technical Indicator Network} (TIN), and the full set of such networks defines the overarching TINs framework. This design enables interpretability, modularity, and adaptive learning, while maintaining semantic fidelity to traditional technical analysis.

The remainder of this paper is organized as follows: Section~2 formalizes the decomposition of classical indicators into modular layer operators. Section~3 presents the general architecture of TINs. Section~4 provides a detailed case study reconstructing indicators within this framework. Section~5 evaluates the proposed architecture through reinforcement learning-driven trading simulations. Section~6 concludes with a discussion of implications and future research directions.

\section{Topological Representation of Technical Indicators in Neural Networks}

Throughout the twentieth century, technical indicators became foundational components of both academic research and practical trading systems. Classical tools such as the Moving Average (MA) and Moving Average Convergence Divergence (MACD) represent only a small subset of a broader class of mathematically defined heuristics that have guided price-based decision-making across generations of market participants. The generation of trading signals is often formalized as:
\begin{equation}
\text{Signal} = F(\text{Market}_t^{(k)})
\end{equation}
where \( F \) denotes a transformation applied to financial data over a specific time window \( k \), producing trading signals at time \( t \). This formulation is widely adopted in quantitative finance. Both MA and MACD compute weighted averages of prices over predefined windows, expressed as:
\begin{equation}
\text{MA}_t^{(k)} = \sum_{i=t-k}^{t} w_i \cdot price_i
\end{equation}

\begin{equation}
\begin{aligned}
    \text{MACD}_t &= \text{MA}_t^{(N^\text{slow})} - \text{MA}_t^{(N^\text{fast})}, \\
    \text{Signal}_t^{(k)} &= \sum_{i=t-k}^{t} w_i \cdot \text{MACD}_i.
\end{aligned}
\end{equation}

\begin{figure}[!t]
    \centering
    \subfloat[Simple Moving Average]{%
        \includegraphics[width=0.48\linewidth]{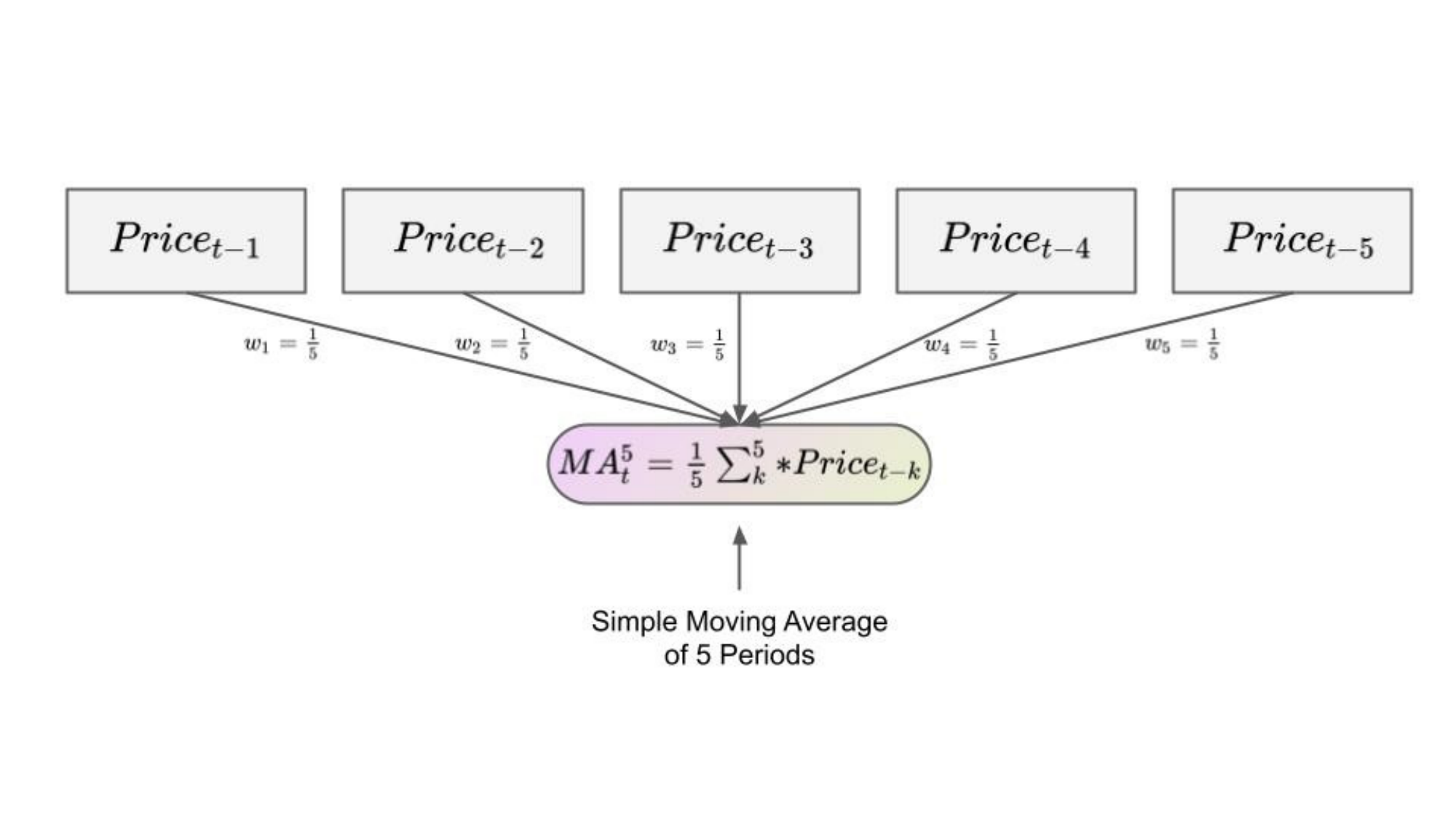}}\hfil
    \subfloat[Exponential Moving Average]{%
        \includegraphics[width=0.48\linewidth]{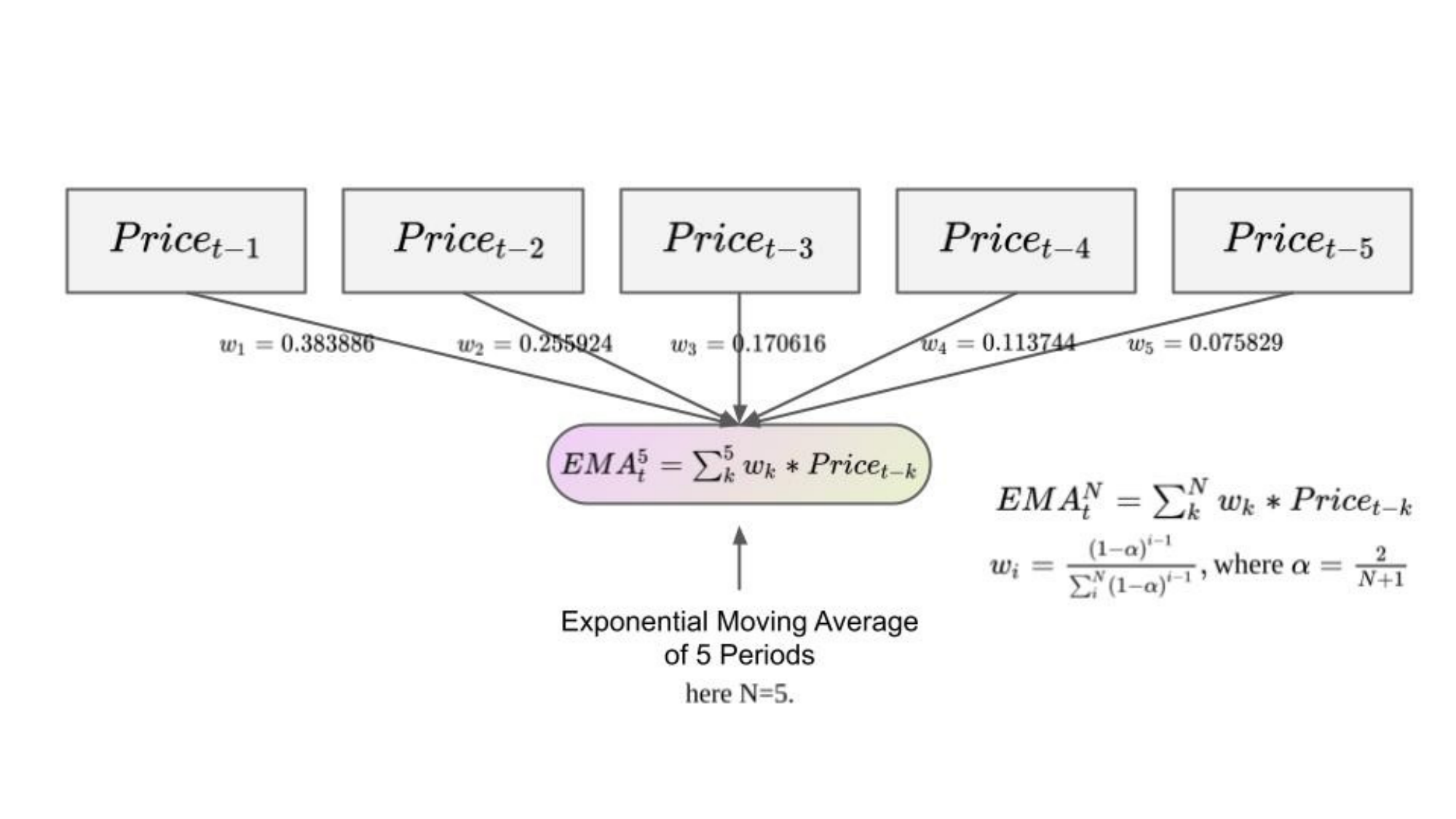}}
    \caption{Examples of Moving Average Indicators in Topological Form.}
    \label{fig:example}
\end{figure}

Equations (1)--(3) reveal a computational form analogous to that of a simple Multilayer Perceptron (MLP), where the weights \( w_i \) are initialized according to the canonical mathematical definitions of the corresponding indicator and encode the temporal weighting schemes. Figure~\ref{fig:example} presents two representative cases—Simple Moving Average (SMA) and Exponential Moving Average (EMA)—each applied over a five-step time window. In both cases, the resulting network topology is isomorphic to a fully connected neural layer with a single output node. Extending this perspective to more complex indicators, the MACD incorporates three distinct time spans—\textit{slow}, \textit{fast}, and \textit{final}—denoted \( N^{type} \), where \( type \in \{\text{slow}, \text{fast}, \text{final}\} \). Its final output is computed by subtracting the two intermediate averages, corresponding to an abstraction operator applied to a pair of vectorized layers. In practical trading applications, the decision signal is generated by the crossover between the MACD line and its signal line. 
More generally, both are expressed as exponential moving averages of the MACD sequence with different spans: 
\[
\text{Signal}_t^{(1)} = \text{MACD}_t, 
\qquad
\text{Signal}_t^{(N^\text{final})} = \text{MA}_t^{(N^\text{final})}(\text{MACD}),
\]
where the case $N=1$ corresponds to the raw MACD line and larger $N^\text{slow}$ (typically $N^\text{slow}=9$) yields the smoothed signal line. 
The crossover can thus be defined as
\[
\text{Crossover}_t = \text{Signal}_t^{(1)} - \text{Signal}_t^{(N^\text{final})}, 
\qquad
\text{TradeSignal}_t = \operatorname{sign}(\text{Crossover}_t),
\]
where $\text{TradeSignal}_t = +1$ indicates a buy signal and $\text{TradeSignal}_t = -1$ a sell signal \citep{kang_improving_2021}.

This structural correspondence is not limited to moving averages. A broad class of widely used technical indicators—including the Commodity Channel Index (CCI), Relative Strength Index (RSI), Stochastic Oscillator, Ultimate Oscillator, and Williams \%R—facilitate each to be decomposed into a finite set of layer operators such as weighted averaging, differencing, pooling, and normalization. This operator-based formulation preserves the original decision logic of the indicators while enabling trainability, robustness, and seamless integration with multidimensional inputs.

\textbf{Definition (Layer Operator Equivalence).}  
A layer operator is equivalent to a technical indicator operation if, under identical parameterization, it yields outputs numerically indistinguishable from the original operation for all admissible inputs, within a prescribed tolerance. Equivalence is defined in terms of computational semantics rather than algebraic identity.

For example, the weighted averaging \circlesym{$\sum$} used in indicators such as the Simple or Exponential Moving Average corresponds directly to a linear layer in deep learning frameworks, where the predefined weights encode temporal contributions. Similarly, arithmetic operations such as addition \circlesym{$+$}, subtraction \circlesym{$-$} , multiplication \circlesym{$\times$}, and division \circlesym{$\div$} are realized as vectorized element-wise transformations within neural layers. Furthermore, rolling operations such as clipping, pooling, mean absolute deviation (MAD), or standard deviation (SD) are reformulated as vectorized operators that take sliding-window arrays as input, rather than single data points as in the original indicator definitions. By mapping these canonical operations into vectorized layer operators, Technical Indicator Networks preserve the semantics of classical indicators while enabling differentiation, parameter learning, and integration with modern deep learning toolchains.

\begin{figure}[!t]
\centering
\includegraphics[width=0.8\linewidth]{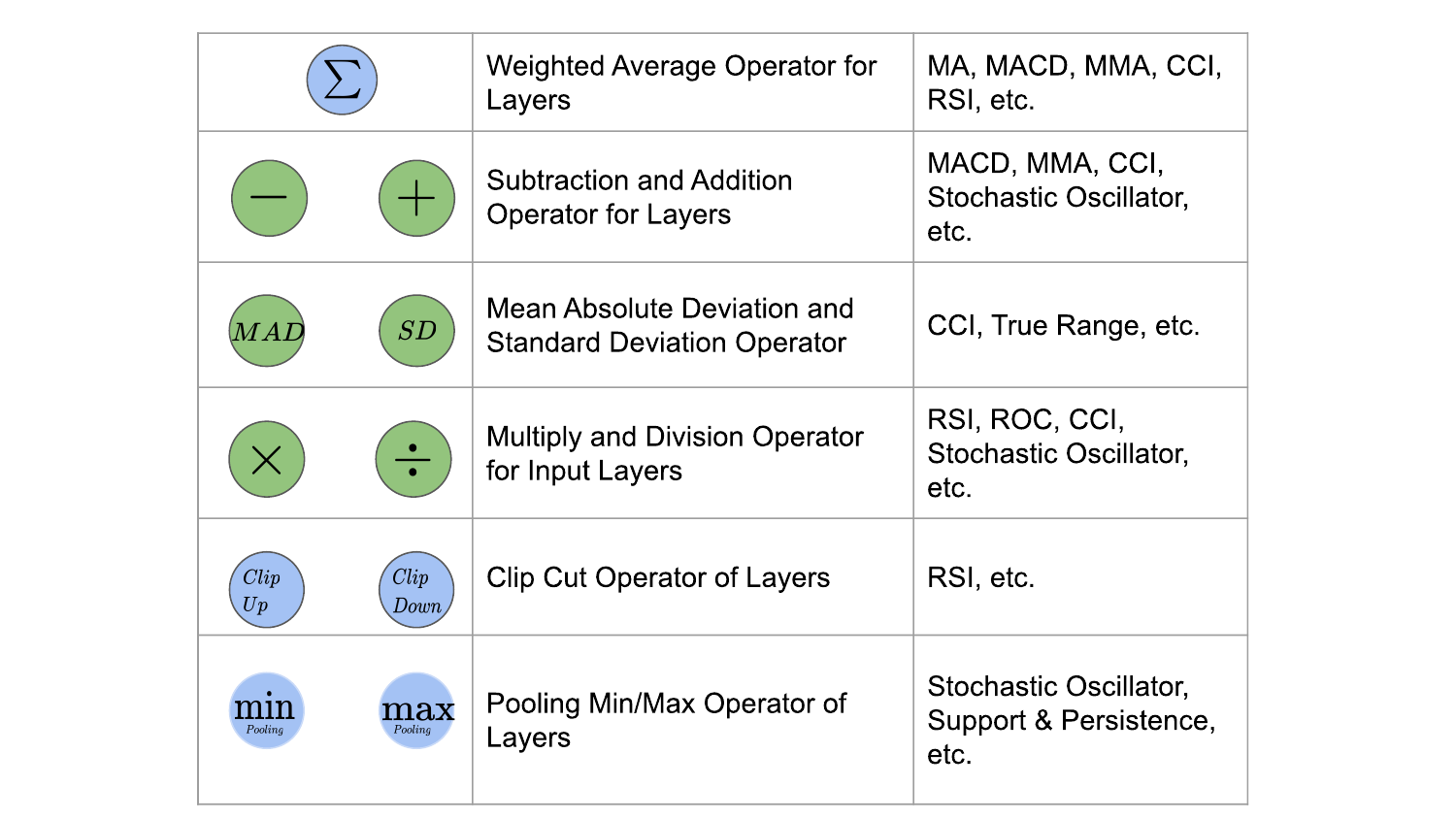}
\caption{Topological Mapping Between Technical Indicators and Neural Network Architectures.}
\label{fig:operator}
\end{figure}

Figure~\ref{fig:operator} presents representative layer operators from major deep learning frameworks (TensorFlow, PyTorch, MXNet) mapped to the technical indicators whose computations they replicate. This one-to-one mapping enables interpretability: practitioners are able to trace model outputs to familiar analytical constructs. To ensure semantic fidelity, the initial parameters of each layer operator are assigned directly from the canonical mathematical definitions of their corresponding indicators. This initialization guarantees that, prior to any learning, the network reproduces exactly the outputs of the original technical indicator. Within TINs, conventional indicators are reinterpreted as trainable topologies composed of these operators, closing the gap between rule-based analysis and data-driven learning. The topology-preserving TINs are complementary to deep reinforcement learning based trading systems and advisor-style models~\cite{Theate2021ESWA_DRL,Felizardo2022ESWA_RL,Majidi2024ESWA_ContinuousRL,Ricchiuti2025ESWA_AdvisorNN}, 
and they naturally support explainability by construction.

With appropriate initialization, a network exactly replicates an indicator signal patterns, preserving operational continuity in live deployment. Weights derived from canonical indicator definitions ensure equivalence prior to training. Subsequent optimization—via supervised or reinforcement learning—permits to adapt parameters to evolving market conditions. In reinforcement learning, a reward function and simulated environment guide parameter updates, enabling the architecture to adjust to non-stationary dynamics while retaining interpretability.

\section{Technical Indicator Networks}

Building upon the conceptual foundation established in the Introduction, the Technical Indicator Networks (TINs) framework is formalized here in terms of its internal architecture, operational semantics, and generalization capabilities. While previously defined as neural realizations of technical indicator logic, this section details how TINs decompose such logic into modular layer operators, enable parameter learning, and extend applicability across diverse financial modalities. By initializing network weights according to canonical indicator definitions, TINs reproduce the functional behavior of classical indicators while supporting multidimensional extensions and adaptive optimization. 

Unlike static indicators governed by fixed mathematical formulas, TINs incorporate reinforcement learning to optimize both structural and operational parameters. This enables dynamic adaptation, allowing the architecture to evolve in response to shifting market regimes and to discover new trading patterns. From a topological perspective, TINs transform conventional univariate indicator structures into multidimensional analytical frameworks. For example, whereas the MACD traditionally operates on a single time series, its TIN-based counterpart integrates heterogeneous inputs such as asset prices, trading volumes, and sentiment-derived feature vectors. This architectural enrichment preserves interpretability while enhancing robustness in cross-market and cross-domain applications.

A review of the literature indicates that no neural network architectures have been explicitly designed for trading applications based on the topological principles of technical indicators. Most deep learning models applied in finance have been adapted from other domains—such as natural language processing or computer vision—driven primarily by performance considerations rather than domain-specific alignment. Although effective in their native contexts, such architectures often lack the structural interpretability and semantic transparency necessary for algorithmic trading. The TINs framework directly addresses this gap by embedding the operational logic of traditional technical indicators within a trainable neural topology, ensuring that every connection and transformation remains traceable to a well-defined financial construct.

Furthermore, TINs are inherently generalizable across instruments, asset classes, and market conditions. For instance, a MACD-based TIN is able dynamically to adjust its internal parameters for equities, commodities, or cryptocurrencies without altering its underlying operator structure. This contrasts with the rigidity of classical MACD implementations, where fixed parameters limit transferability. As shown in Figure~\ref{fig:reflection2}, the TIN approach supports concurrent processing of heterogeneous data sources—price series, volumes, sentiment signals—within a unified indicator topology. This multidimensional capability, combined with topology-preserving initialization and subsequent trainable adaptation, enhances predictive performance and supports the generation of actionable insights under diverse trading conditions.

\begin{figure}[!t]
\end{figure}
\begin{figure}[!t]
    \centering
    \subfloat[Technical Indicator Network representation of MMA]{%
        \includegraphics[width=0.48\linewidth]{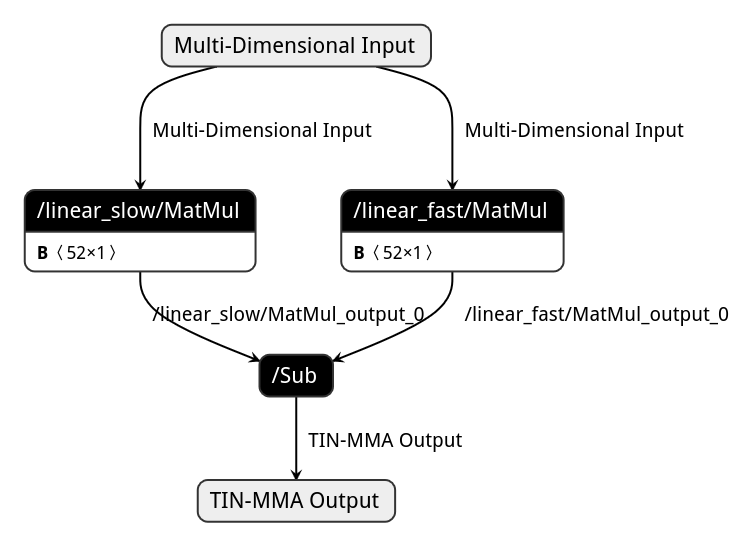}}\hfil
    \subfloat[Technical Indicator Network representation of MACD]{%
        \includegraphics[width=0.48\linewidth]{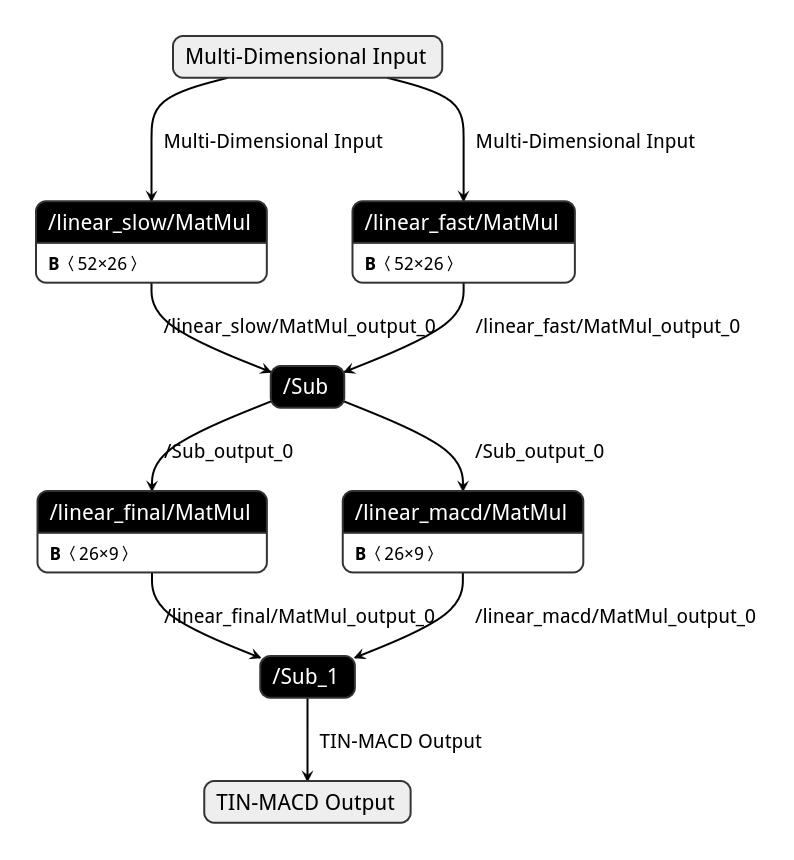}}
    \caption{Pytorch implementation of TINs for MMA and MACD Indicators.}
    \label{fig:visualization}
\end{figure}

\section{Case Study: Reconstruction of MACD into a Technical Indicator Network}

Brokers frequently provide extensive libraries of technical indicators, with widely available toolkits (e.g., TA-Lib) implementing more than one hundred indicators in multiple programming languages. The TINs framework reconstructs such indicators by initializing directly from their canonical mathematical definitions, mapping each computation to a corresponding layer operator, and preserving the original topology while enabling trainable adaptation. In contrast to these static implementations, the TINs framework offers a modern, adaptable, and interpretable neural realization of such indicators, implemented directly in Python-based deep learning environments. This approach ensures accessibility for both academic research and industrial deployment. Figure~\ref{fig:visualization} illustrates the TINs implementation of two representative indicators, Multiple MA and MACD using the PyTorch framework. The network topology is defined in terms of layer operators, with weights initialized according to the canonical definitions of each indicator. This design preserves the original indicator logic while enabling subsequent parameter optimization through reinforcement learning or other adaptive methods.

In this section, the Moving Average Convergence Divergence (MACD) indicator—referenced in \cite{kang_improving_2021}—is reconstructed within the TINs framework. By employing linear layers with predefined layer operators and initializing their weights to represent fixed-period computations for the slow and fast moving averages, the TIN reproduces the operational logic of the MACD. Moreover, the architecture extends MACD from a one-dimensional formulation into a multidimensional topology, enabling the integration of cross-market signals, sentiment-derived vectors, and other auxiliary features. This enhancement increases adaptability and scalability, thereby addressing the complexity of modern financial environments.

\subsection{Basic MACD Layer}

\begin{figure}[!t]
\centering
\includegraphics[width=0.86\linewidth]{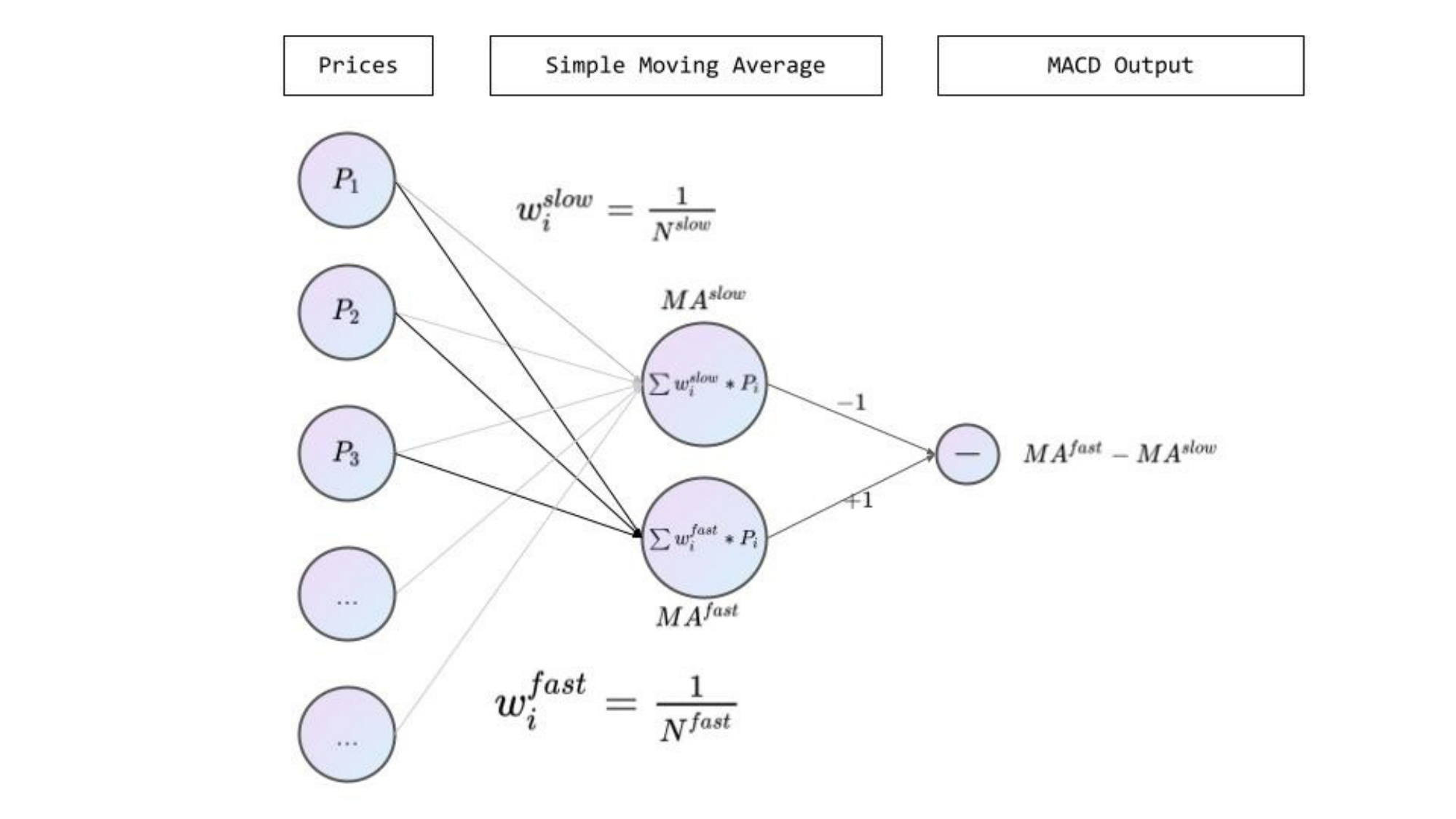}
\caption{Technical Indicator Network representation of MACD.}
\label{fig:maNN}
\end{figure}

As shown in Figure~\ref{fig:maNN}, the network topology derived from the moving average operator functions as a linear layer without non-linear activation, consistent with the description in Section~2. It processes input sequences by separately computing the fast and slow moving averages through two parallel linear layers. These outputs are combined using a subtraction layer operator, denoted as $\ominus$, to produce the differential signal. 

By configuring the edge weights according to established moving average definitions, the TIN reproduces both simple and exponential moving average strategies without requiring initial data-driven training. This structure generalizes a broad class of MA-based systems and remains interpretable. Increasing the number of hidden nodes augments analytical capacity. For the Simple Moving Average (SMA), edge weights are initialized as
\[
w_{i}^{type} = \frac{1}{N^{type}}, \quad type \in \{\text{slow}, \text{fast}\},
\]
over a fixed time window of the price series. This initialization forms the baseline for reinforcement learning–based optimization while preserving the behaviour of conventional SMA strategies.

For the Exponential Moving Average (EMA), weights are initialized as
\[
w_{i}^{type} = \frac{(1-\alpha)}{\sum_{t=1}^{N^{type}}(1+\alpha)^{t-1}}, \quad \alpha = \frac{2}{N^{type}+1}.
\]
Under this initialization, the TIN outputs remain consistent with EMA-based strategies, providing a theoretically aligned starting point for reinforcement learning–driven refinement. In this way, the MACD-based Indicator Network is strictly initialized from its classical mathematical formulation, ensuring that its pre-training behavior is indistinguishable from the traditional indicator before subsequent reinforcement learning optimization.

\subsection{MACD Network}

\begin{figure}[!t]
\centering
\includegraphics[width=\linewidth]{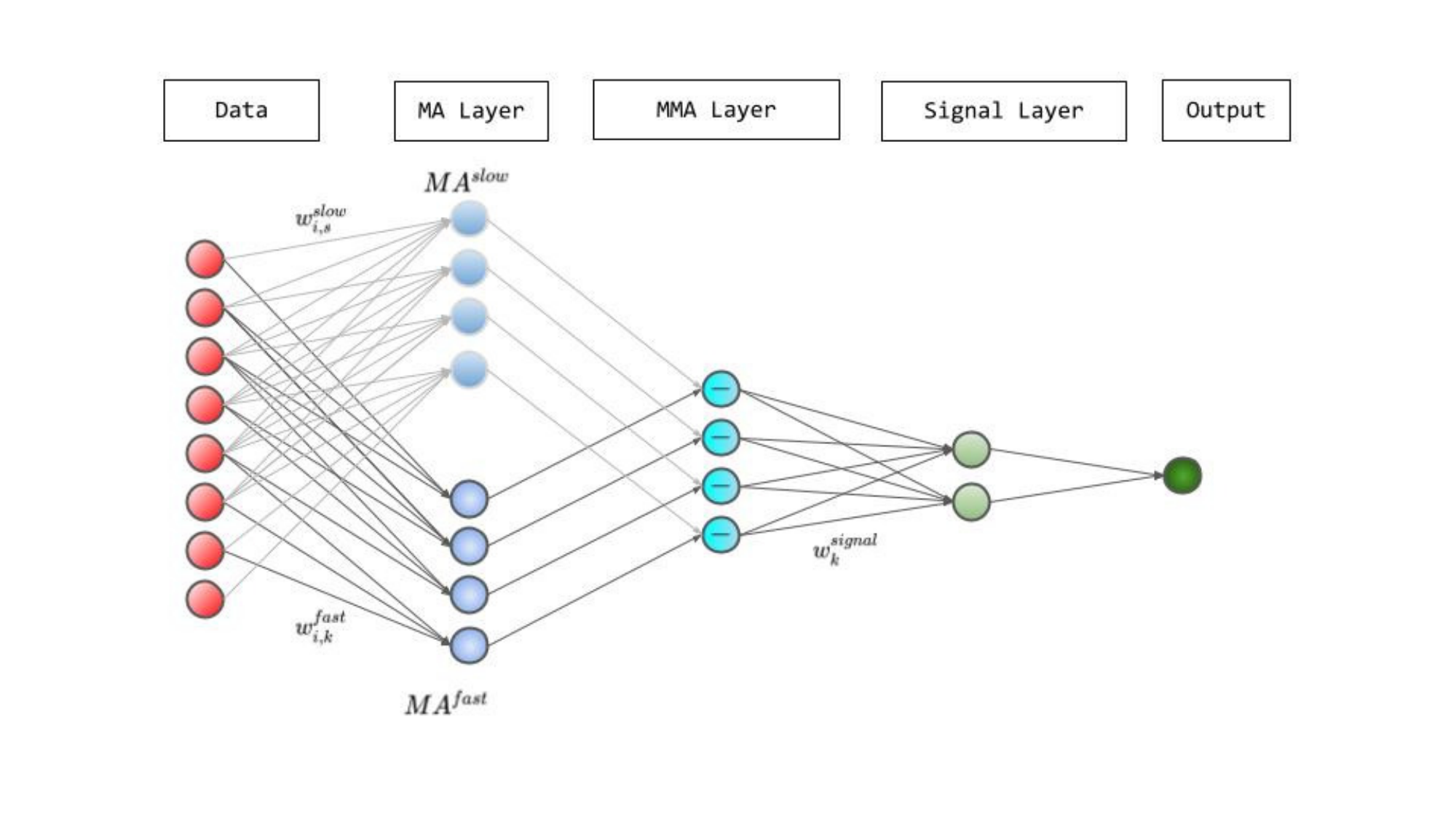}
\caption{Technical Indicator Network topology of MACD oscillator for algorithmic trading.}
\label{fig:macdNN}
\end{figure}

Figure~\ref{fig:macdNN} presents the complete TIN topology for replicating and extending a MACD oscillator. The input layer encodes historical price data or other relevant time series, which are processed in parallel through the fast and slow moving average branches. Their outputs pass through a MACD layer that applies a differential operator, followed by a smoothing layer for the signal line. The final output node generates the MACD oscillator value. 

The network connections and initial weights are configured to reproduce the canonical MACD formulas while allowing for the inclusion of auxiliary inputs. When initialized according to Section~4.1, this architecture reproduces the behaviour of a classical MACD system (see also Figure~\ref{fig:example}). At the same time, its modular design supports seamless integration with additional data streams.

\begin{figure}[!t]
    \centering
    \includegraphics[width=0.8\linewidth]{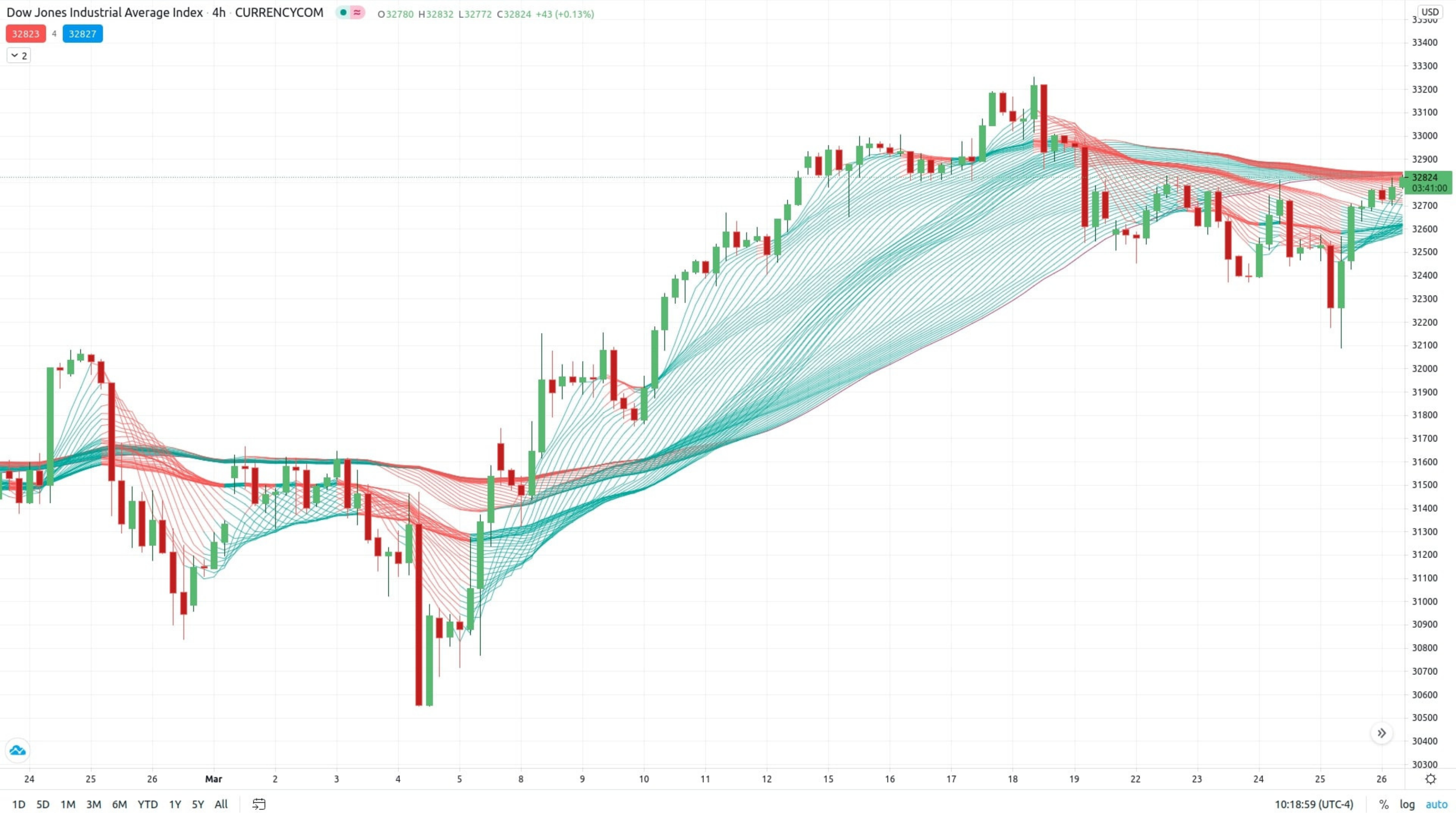}
    \caption{Technical Indicator Network: multiple moving average trading platform in practice.}
    \label{fig:reflection1}
\end{figure}

A principal advantage of this formulation lies in its transparency. Unlike opaque architectures such as LSTM, CNN, or transformer-based models, each node and connection in the MACD TIN corresponds to a well-defined financial formula. This allows practitioners to trace signal transformations step by step. Figure~\ref{fig:reflection1} shows the corresponding EMA overlays on price data, confirming fidelity to the original indicator logic.

\subsection{MACD Technical Indicator Network for Multidimensional Inputs}

\begin{figure}[!t]
    \centering
    \includegraphics[width=0.85\linewidth]{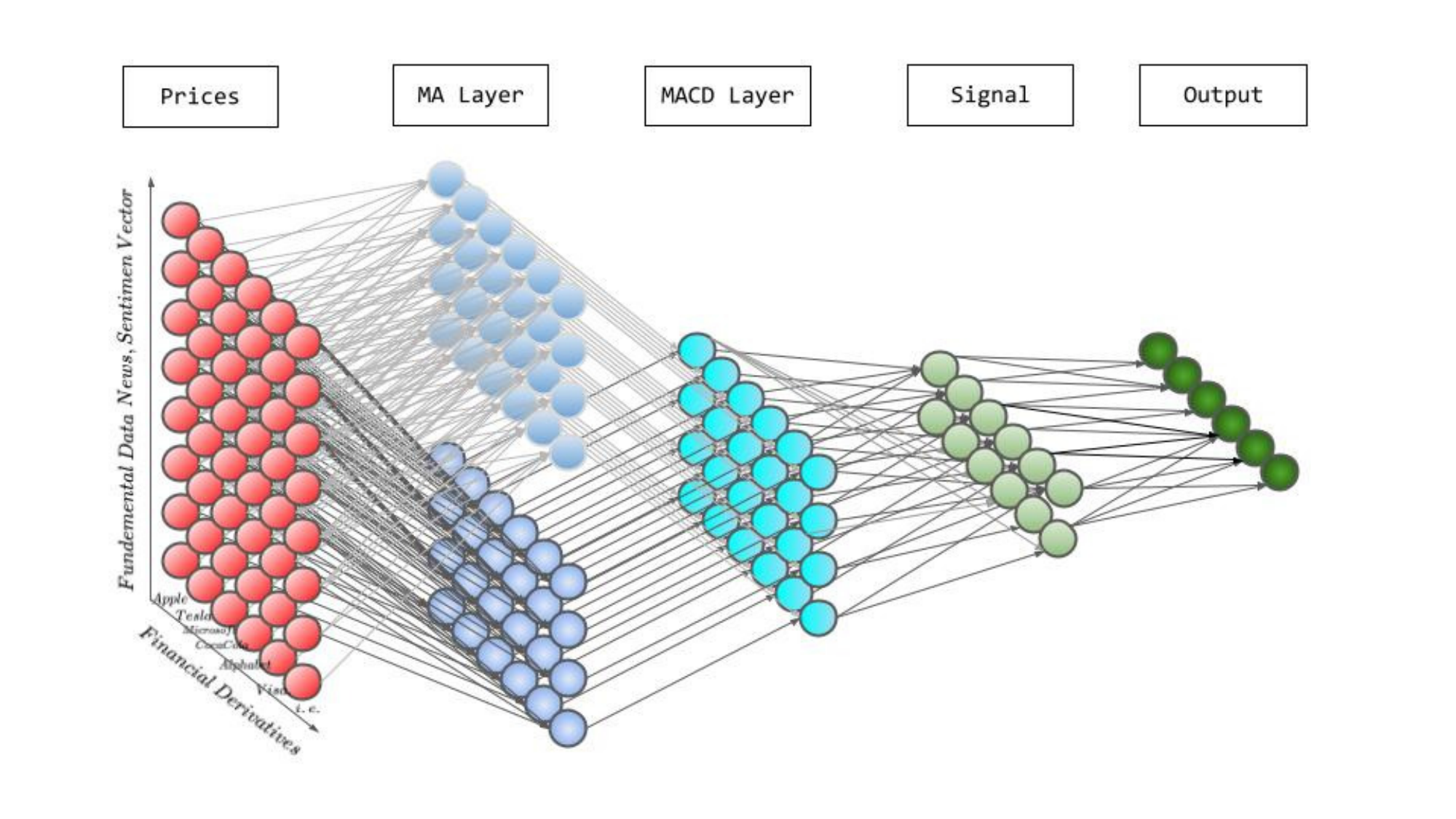}
    \caption{Topology of a multidimensional moving average indicator network.}
    \label{fig:macd_net_nD}
\end{figure}

Conventional MACD indicators are typically applied to univariate price series, as in Figure~\ref{fig:reflection1}. This restriction limits their ability to capture the interconnected and multi-modal dynamics of modern financial markets. Within the TIN framework, the MACD operator is generalized to process multidimensional inputs, including sentiment-derived vectors, cross-market prices, and macroeconomic indicators. As shown in Figure~\ref{fig:macd_net_nD}, the same layer operator structure is preserved, enabling the direct incorporation of heterogeneous data types while maintaining the core computational logic of the original indicator. This extension supports modelling of market interdependencies, temporal relationships, and non-price drivers in a unified analytical pipeline.

By reparameterizing traditional indicators such as MA and MACD into structured sequences of layer operators, the TINs framework recasts classical technical analysis as a neural formalism tailored for AI-driven algorithmic trading. The construction principles demonstrated here are readily applicable to other indicators. For instance, implementing the Commodity Channel Index (CCI) within a TIN entails applying pooling and clipping operators over high and low prices, followed by a normalized division operator. Detailed mappings between classical indicators and their corresponding layer operators are provided in the Appendix.

\begin{figure}[!t]
    \centering
    \includegraphics[width=0.8\linewidth]{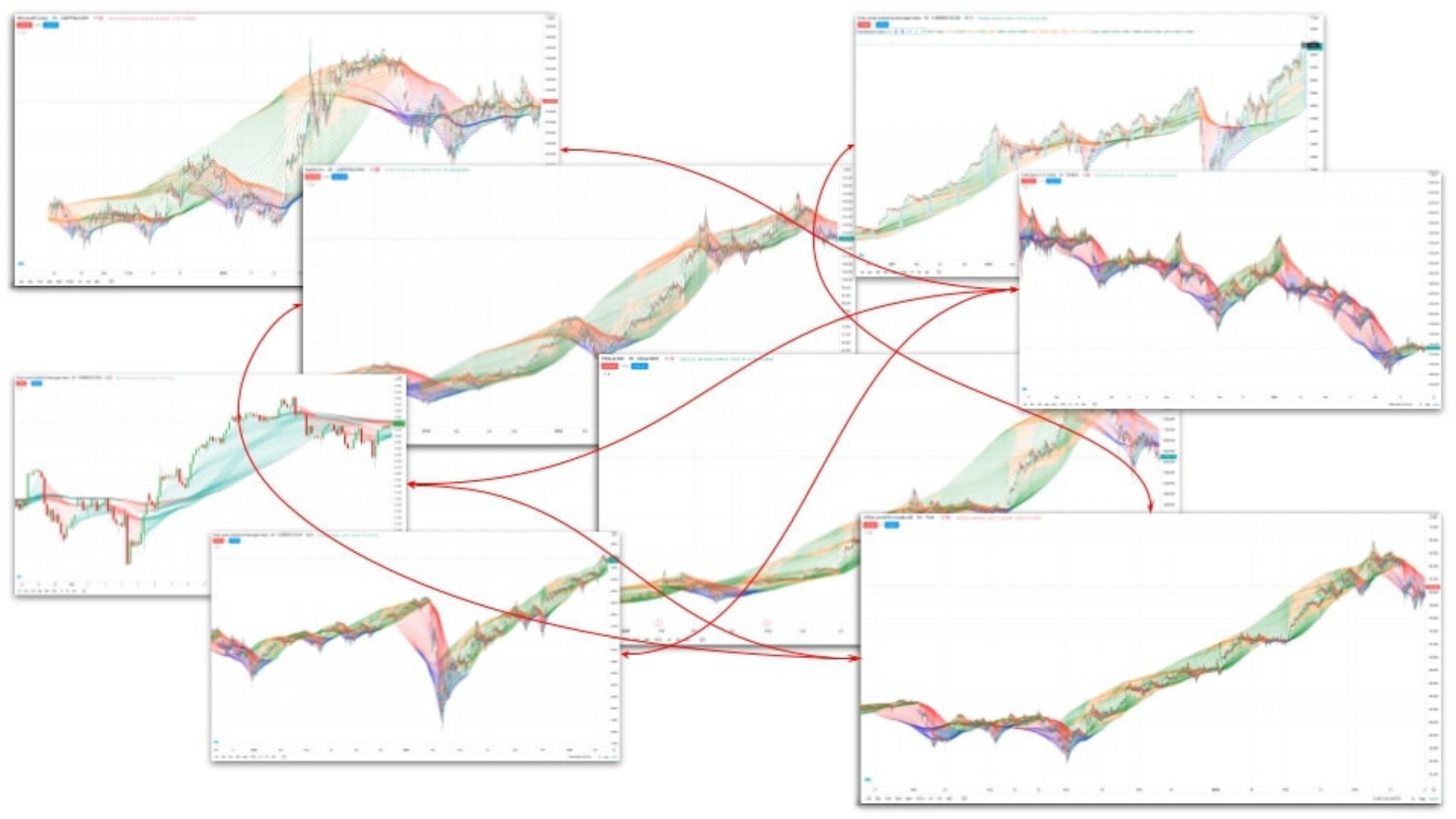}
    \caption{As the next generation of indicators, Technical Indicator Networks create a new market paradigm by upgrading trading platforms with cross-market visibility and enhanced capacity to deliver more efficient information for end-users. While no comparable platform currently exists, this functionality is expected to become a natural and essential feature of future trading systems, underscoring the significant commercial potential of this approach.}
    \label{fig:reflection2}
\end{figure}

Figure~\ref{fig:reflection2} illustrates the capacity of TINs to integrate complex relationships among heterogeneous input sources and cross-market dynamics within a single unified model, demonstrates how TINs enable multiple moving average lines for multiple stocks with cross-connections. This capability marks a substantial advantage over traditional indicators, which are generally restricted to single-asset analysis and lack the flexibility to incorporate context-rich signals. Technical Indicator Networks address these limitations by enabling multi-asset interactions and multi-modal signal fusion within a unified architectural framework, while preserving the canonical topology of the original indicator through mathematically grounded initialization.

\section{Performance Test of MACD TIN}

To evaluate the practical viability and adaptive potential of the proposed framework, a performance assessment is conducted on a MACD-based Indicator Network (TIN-MACD), initialized from the canonical MACD mathematical definition and expressed as a topology-preserving layer-operator network, designed to replicate, and potentially enhance, the trading logic embedded in the traditional MACD indicator. The selected TIN configuration comprises networks strictly initialized from the canonical mathematical definition of MACD via layer operators, ensuring that pre-training outputs are identical to the classical formulation. Layer operators and hyperparameters directly reflect the original specification, while reinforcement learning as in \cite{Deng2017TNNLS_DeepDirectRL} is applied to optimize parameters in a simulated trading environment. Specifically, the policy network $\pi_\theta(a_t|s_t)$, which is TIN-MACD in this case study, parameterized by $\theta$, is updated using the reinforce gradient \citep{Williams1992REINFORCE}:
\[
\nabla_\theta J(\theta) \approx \sum_t \nabla_\theta \log \pi_\theta(a_t|s_t)\,(G_t - b),
\]
with discounted return $G_t = \sum_{k=t}^T \gamma^{k-t} r_k$ and baseline $b$ for variance reduction. The parameters are adjusted by stochastic gradient ascent,
\[
\theta \leftarrow \theta + \alpha \sum_t \nabla_\theta \log \pi_\theta(a_t|s_t)\,(G_t - b),
\]
with learning rate $\alpha>0$, thereby reinforcing profitable trading decisions. This enables dynamic adaptation to non-stationary market conditions while preserving the indicator topological structure. In the implementation, the discounted returns $G_t$ are normalized within each batch, i.e., $adv_t = (G_t - \bar{G}) / \sigma_G$, 
with $\bar{G}$ the batch mean and $\sigma_G$ the standard deviation. 
This corresponds to using the batch mean return as baseline \citep{Williams1992REINFORCE} with an additional variance normalization step, 
a practice commonly adopted in modern reinforcement learning \citep{Schulman2015GAE,Mnih2016A3C}. In addition, to enhance sample efficiency and stability, experience replay \citep{Mnih2015DQN} is adapted, 
where recent trajectories are retained in a sliding window and resampled for policy updates.

Each network processes a 52-day rolling window of historical prices through a hidden layer of 26 units, consistent with the MACD parameterization $(12,26,9)$. Trading actions, buy or sell, as MACD crossover trading signal defintion, are generated from network outputs and executed using adjusted closing prices to ensure simulation fidelity. Risk-adjusted performance is measured via Sharpe and Sortino ratios, computed on adjusted daily returns, consistent with prevailing practice in quantitative trading evaluation. Transaction costs were not considered, since this work is intended as a proof-of-concept demonstration. Unlike prior AI-trading works that often validate on a single product, this study covers the full DJIA 30 constituents, which is already substantially broader for a proof-of-concept. The evaluation follows a fair progression: the canonical MACD serves as the baseline, followed by its topology-equivalent TIN initialized directly from the canonical formulation, and finally the trainable TIN variants, including both the price-only configuration and the extended configuration incorporating On-Balance Volume (OBV). This design isolates the contribution of topology preservation and adaptive training without confounds from heterogeneous model classes.

It is important to emphasize that direct comparisons with sequence models such as LSTM, RNN, or Transformer are fundamentally inappropriate in this context. These architectures inherently treat indicators as static input features, which is precisely the limitation that TINs are designed to overcome. Feeding TIN outputs or indciators into an LSTM or RNN no longer constitutes a comparison against the original indicator but instead creates a hybrid system, obscuring the proof-of-concept validation. Empirical evidence shows that TIN-MACD with price and OBV already achieves statistically supported improvements within a simple MLP architecture, which naturally extends to deeper neural structures. This provides a partial form of cross-model validation without undermining the conceptual contribution of TINs. 

\begin{figure}[!t]
    \centering
    \subfloat[Traditional MACD (12,26,9)]{%
        \includegraphics[width=0.45\linewidth]{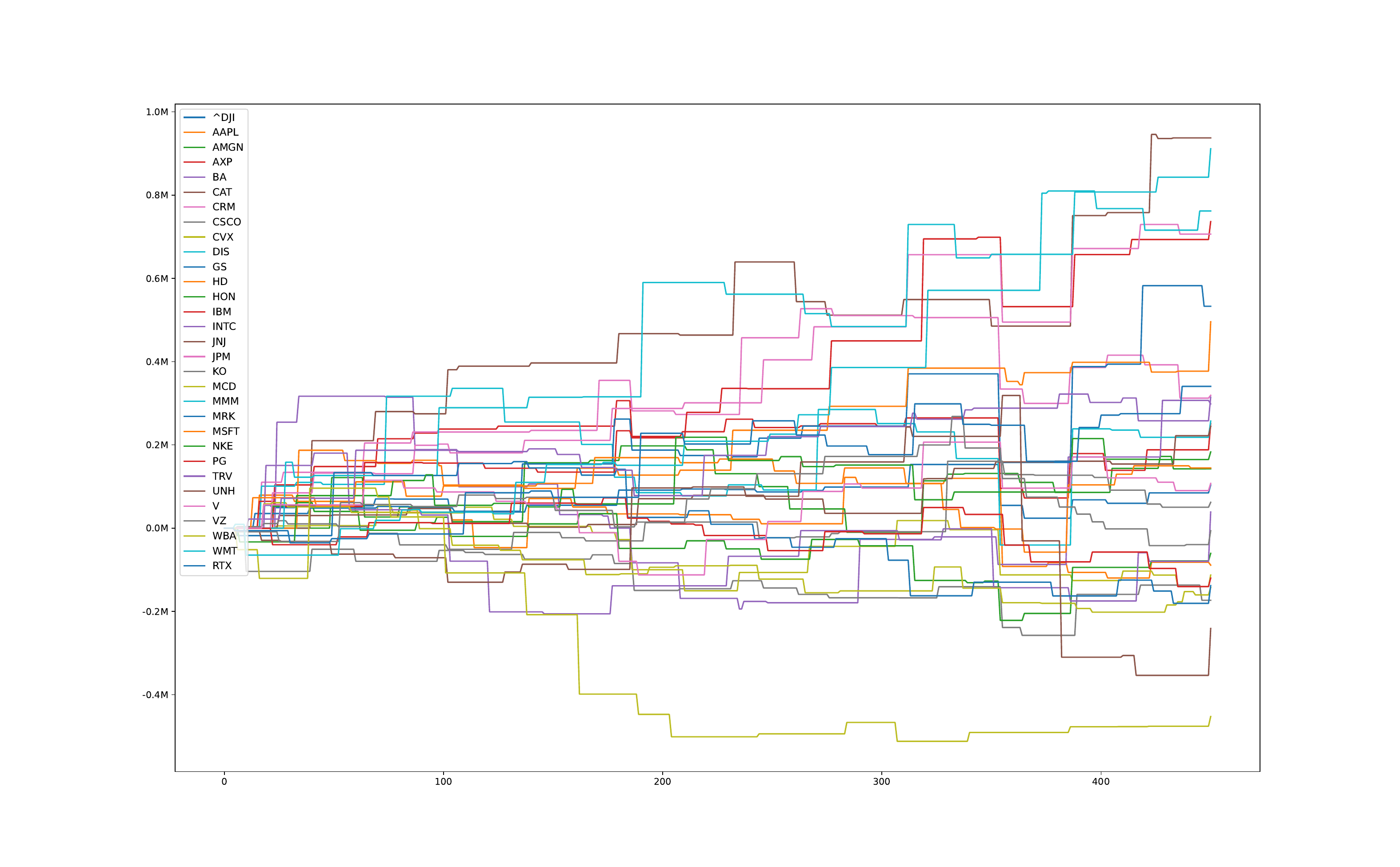}}\hfil
    \subfloat[TIN-MACD (Price) with 52-day windows and 26 hidden units]{%
        \includegraphics[width=0.45\linewidth]{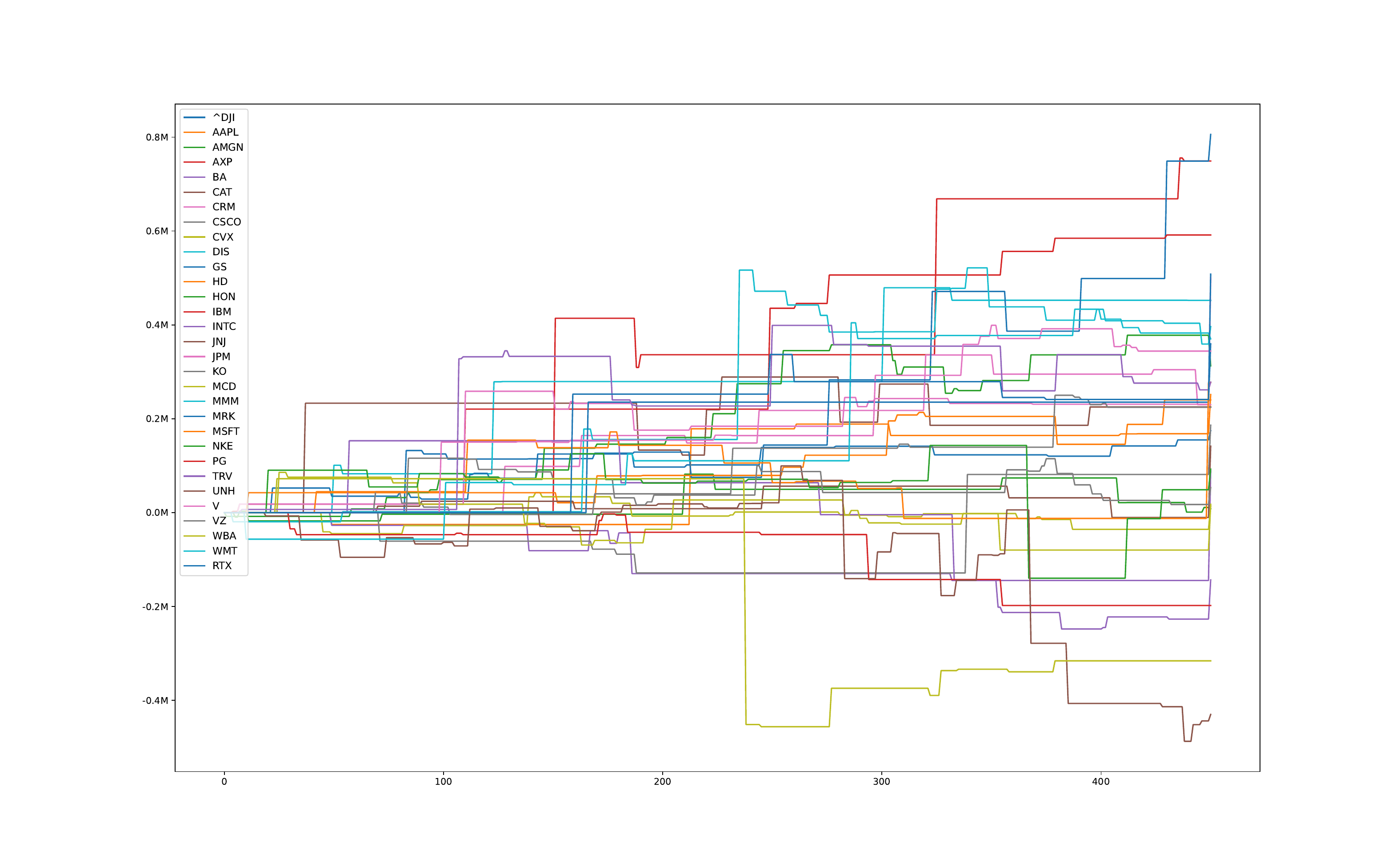}}
    \par\medskip
    \subfloat[TIN-MACD (Price + OBV) with 52-day windows and 26 hidden units]{%
        \includegraphics[width=0.95\linewidth]{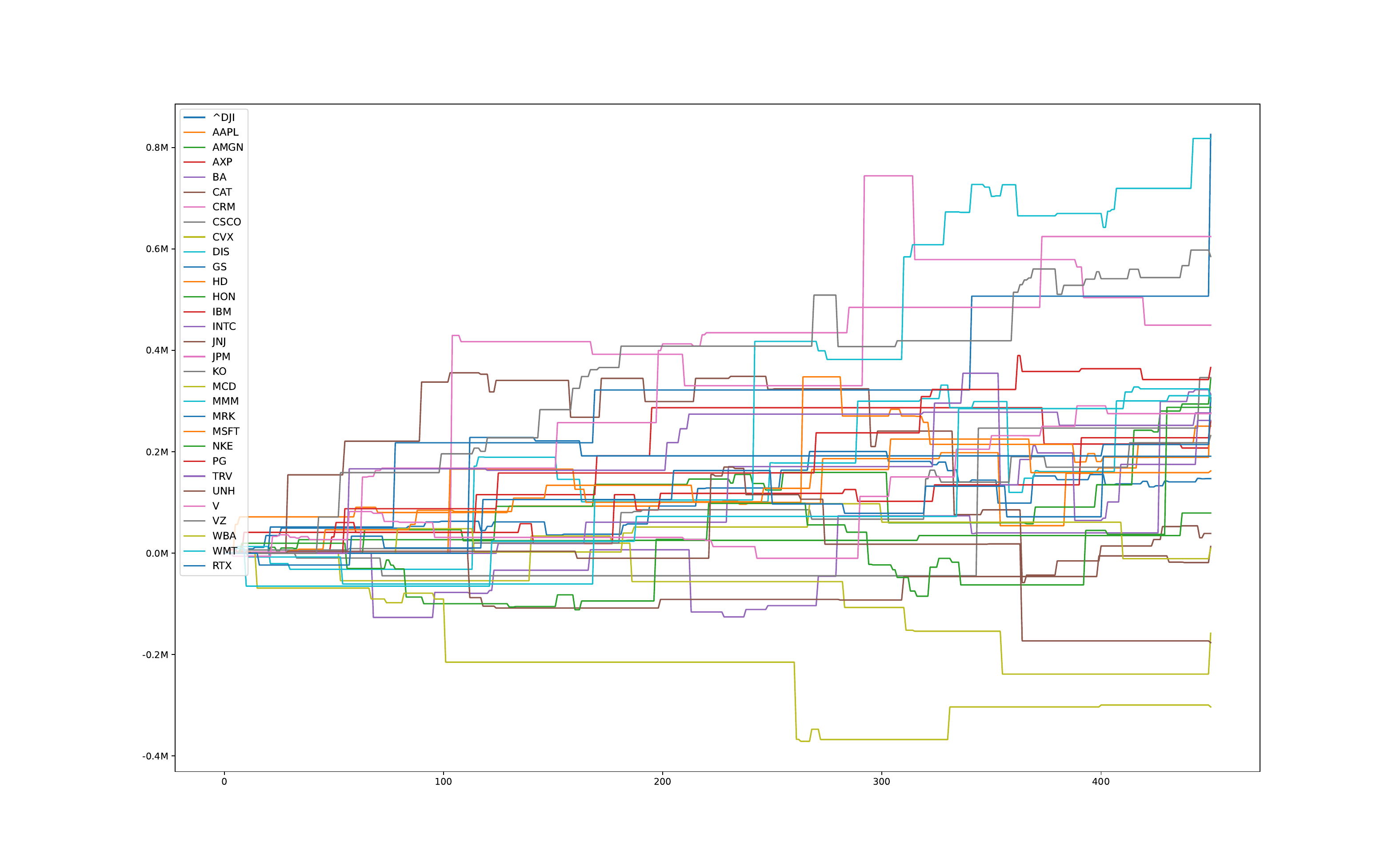}}
    \caption{Out-of-sample performance evaluation on US30 components across a three-year testing horizon.}
    \label{fig:us30c}
\end{figure}

The dataset comprises daily adjusted closing prices for the 30 constituents of the US30 index, retrieved via the publicly available \texttt{yfinance} library. The sample is divided into an in-sample training period from 2015-08-25 to 2023-08-21 and an out-of-sample evaluation period from 2023-08-22 to 2025-08-22. Each equity is modeled using an independent TIN-MACD instantiation, with identical hyperparameter settings applied to ensure comparability across constituents. To examine the impact of incorporating volume-based information, On-Balance Volume (OBV) is computed as the cumulative daily trading volume adjusted by the sign of price movements and introduced as an auxiliary input in selected TIN-MACD variants. Figures~\ref{fig:us30c} (a) and (b) present a visual comparison between the traditional MACD $(12,26,9)$ and the TIN-MACD implementation. Both strategies operate at relatively low trading frequencies—typically with holding periods of several days to weeks—yet their return distributions differ substantially. Whereas the traditional MACD generates a mixed profile with a few strong performers but many average cases, the TIN-MACD formulation exhibits a positively skewed distribution, indicating that a larger proportion of equities derive measurable gains under the TIN architecture. At the portfolio level, the equally weighted average returns across all 30 constituents further demonstrate that both TIN-MACD configurations (price-only and OBV-augmented) achieve higher cumulative returns than the canonical MACD, while also outperforming the buy-and-hold baseline of the US30 index in terms of risk-adjusted measures.

\begin{table}[!t]
\centering
\caption{Overall Performance Comparison}
\label{tab:overall_performance}
\scriptsize 
\setlength{\tabcolsep}{2pt} 
\renewcommand{\arraystretch}{1.15} 
\begin{tabular*}{\linewidth}{@{\extracolsep{\fill}}lcccc@{}}
\toprule
\textbf{Metric} &
\begin{tabular}[c]{@{}c@{}}\textbf{TIN-MACD}\\(Price, OBV)\end{tabular} &
\begin{tabular}[c]{@{}c@{}}\textbf{TIN-MACD}\\(Price Only)\end{tabular} &
\textbf{Traditional MACD} &
\begin{tabular}[c]{@{}c@{}}\textbf{US30}\\\textbf{Index}\end{tabular} \\
\midrule
Sharpe Ratio & \textbf{3.5323} & 2.3387 & 1.8461 & 1.1236 \\
Sortino Ratio & \textbf{4.7333} & 4.4620 & 1.9680 & 1.4889 \\
Cumulative Return & 0.2309 & 0.1926 & 0.1931 & \textbf{0.3461} \\
\bottomrule
\end{tabular*}
\end{table}

\begin{figure}[!t]
    \centering
    \includegraphics[width=0.99\linewidth]{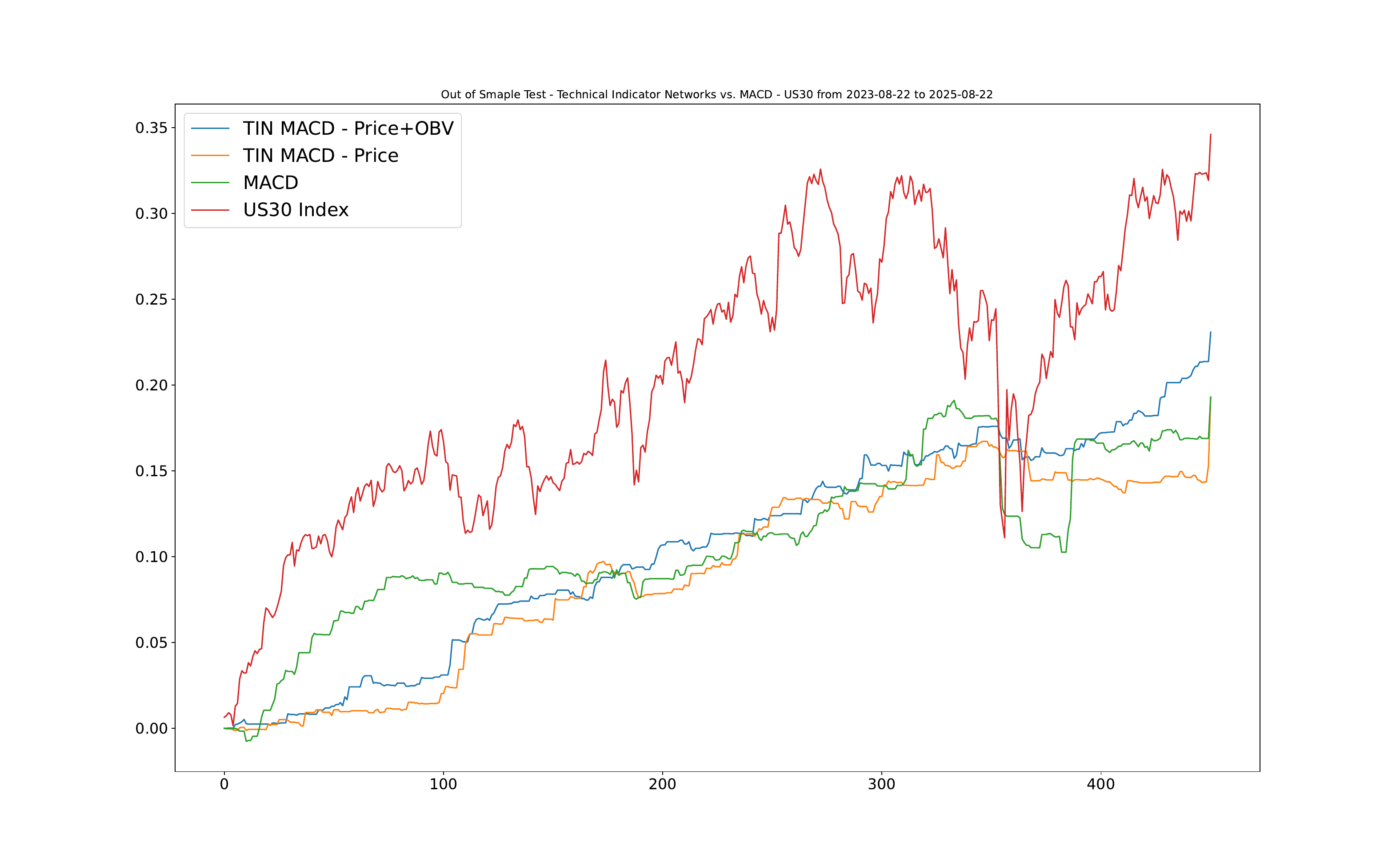}
    \caption{Performance Comparison: TIN-MACD vs.\ traditional MACD on US30 Components.}
    \label{fig:performance}
\end{figure}

Table~\ref{tab:overall_performance} summarizes the aggregate performance metrics, including Sharpe and Sortino ratios as well as cumulative returns. Both TIN-MACD configurations exceed the traditional MACD on risk-adjusted criteria, with the OBV-augmented variant delivering the strongest improvements. Although the US30 index buy-and-hold benchmark achieves the highest absolute cumulative return, the TIN-MACD strategies demonstrate superior Sharpe and Sortino profiles, underscoring their advantage in balancing return with risk. These results highlight that cumulative and risk-adjusted performance measures lead to different conclusions, reinforcing the necessity of Sharpe and Sortino ratios for fair strategy evaluation. Figure~\ref{fig:performance} illustrates the cross-sectional distribution of Sharpe and Sortino ratios across the 30 constituents. For detailed statistical tests and robustness checks, readers are referred to \ref{sec:tin_us30_performance}.

\section{Summary}

This manuscript introduced Technical Indicator Networks (TINs), a novel framework that reformulates classical technical indicators into topology-preserving, trainable neural architectures. By initializing from canonical indicator definitions and expressing their operations as layer operators, TINs retain the semantic logic of traditional heuristics while enabling parameter refinement in trading-specific contexts. The modular design integrates with advanced neural architectures and LLM-guided agents, supporting scalable, interpretable, and adaptable trading systems.

The experiments on MACD-based TINs across the 30 constituents of the US30 index were designed as a proof-of-concept to validate feasibility rather than to maximize performance. The empirical evaluation confirmed the expected performance ordering: TIN-MACD with Price+OBV consistently outperformed the price-only configuration, which itself exceeded the canonical MACD. On risk-adjusted metrics, Sharpe ratio improvements were statistically significant in the OBV-augmented variant and borderline significant in the price-only variant, while Sortino ratio enhancements were positive on average but less consistent across constituents. These findings indicate that the principal contribution of TINs lies in bridging traditional heuristics with neural learning through topology-preserving architectures, enabling interpretable extensions of classical strategies rather than short-term performance maximization.

Although limited to simulated trading on the US30 due to resource constraints, the framework is readily extensible to other indicators, asset classes, and deployment environments. Future research directions include integration with advanced neural architectures and LLM-generated agents, incorporation of broker-side APIs for real-time deployment, enrichment of input modalities, enhancement of reinforcement learning protocols, and benchmarking within realistic market simulators.

In addition to its theoretical significance, the proposed framework carries substantial commercial implications. As the next generation of indicators, Technical Indicator Networks create the foundation for upgrading trading platforms with cross-market visibility and enhanced decision-support capacity. Although no such platforms currently exist, this functionality is expected to become an essential feature of future trading systems, underscoring the significant commercial potential of this approach in shaping the next generation of financial technology.

In conclusion, TINs provide a generalizable foundation for interpretable, adaptable, and risk-aware AI models in trading, delivering robust improvements in risk-adjusted performance and offering both academic relevance and practical applicability. As the inaugural study in this direction, the present work contributes primarily by establishing the theoretical basis and demonstrating feasibility through proof-of-concept experiments. Future research is expected to expand the framework with larger-scale empirical validation and systematic comparisons.

\section*{Author biography}

Longfei Lu is a principal data scientist with extensive experience in machine learning and artificial intelligence for quantitative finance and algorithmic trading. He has held data science roles in the financial industry, including at Deutsche Bank, where he works on data-driven decision support, model development, and the deployment of AI systems in production environments. His academic background includes studies at the University of Bremen, and he holds the Principal Data Scientist (PDS\texttrademark) credential awarded by the Data Science Council of America. His research interests focus on interpretable neural architectures for financial decision support, technical-indicator-based neural models, neural-symbolic integration, and risk-aware learning methods for trading systems.

\appendix

\section{Moving Average (MA) and Moving Average Convergence Divergence (MACD) Indicator Networks}

\begin{figure}[!t]
\centering
\includegraphics[width=0.98\linewidth]{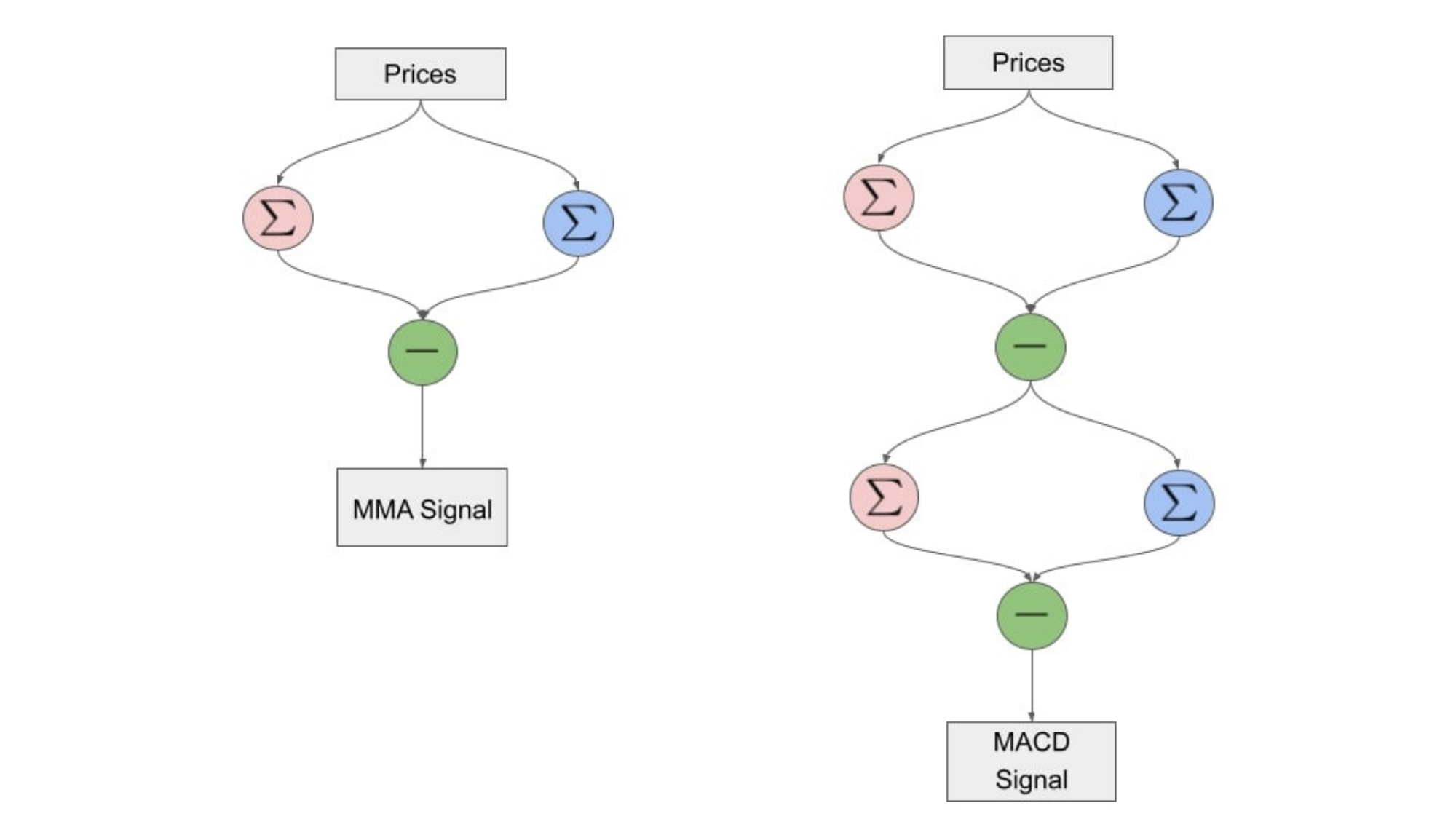}
\caption{Topology of Indicator Network Expressing MA and MACD.}
\label{fig:macdTopology}
\end{figure}

Within the Technical Indicator Networks (TINs) framework, the Moving Average Convergence Divergence (MACD) indicator is reformulated as a modular composition of layer operators computationally equivalent to the operations in its canonical formulation. In accordance with the \textit{Layer Operator Equivalence} principle, each operator is initialized directly from the mathematical definition of the corresponding computation, thereby preserving the semantic integrity of the original indicator while embedding it within a trainable neural topology.

The MACD TIN consists of two sequential smoothing modules—each with a distinct lookback period—implemented as differentiable weighted-average layers. When initialized with canonical exponential moving average (EMA) weights, these layers produce outputs identical to the traditional EMA calculation. Their outputs are passed to an element-wise subtraction operator, measuring the momentum differential between the fast and slow averages, followed by a secondary weighted-average operator to generate the MACD signal line. This construction reproduces the full MACD computation while enabling adaptive weight adjustment through learning algorithms, combining topology preservation, mathematical fidelity, and neural trainability.

\section{Relative Strength Index (RSI) and Rate of Change (ROC) Indicator Networks}

\begin{figure}[!t]
\centering
\includegraphics[width=0.88\linewidth]{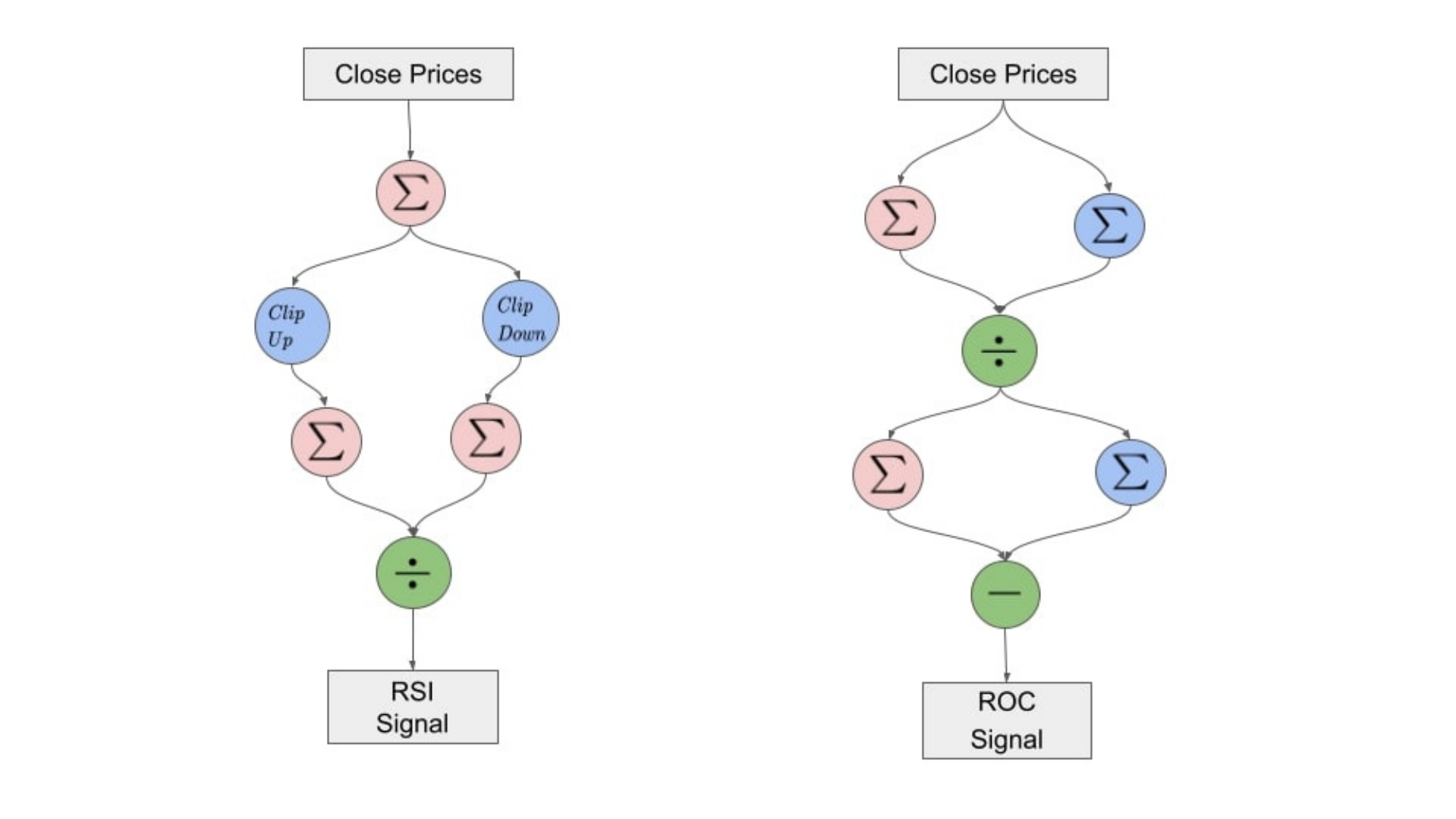}
\caption{Topology of Indicator Network Expressing RSI and ROC.}
\label{fig:rsi_roc}
\end{figure}

Within the TINs framework, the Relative Strength Index (RSI) and Rate of Change (ROC) indicators are expressed as modular compositions of division layer operators computationally equivalent to their classical momentum-based definitions. Following the \textit{Layer Operator Equivalence} principle, each operator is initialized from its canonical formula, ensuring semantic preservation while enabling integration into a trainable neural architecture.

The RSI module computes the ratio of smoothed gains to smoothed losses over a defined lookback period, while the ROC module evaluates proportional price changes over a specified time window. Both computations are implemented in vectorized neural modules, with denominator-regularized division operators to ensure numerical stability and gradient continuity. This preserves the ratio-based momentum interpretation while enabling learning-driven adaptation.

\section{Stochastic Oscillator Indicator Network}

\begin{figure}[!t]
\centering
\includegraphics[width=0.98\linewidth]{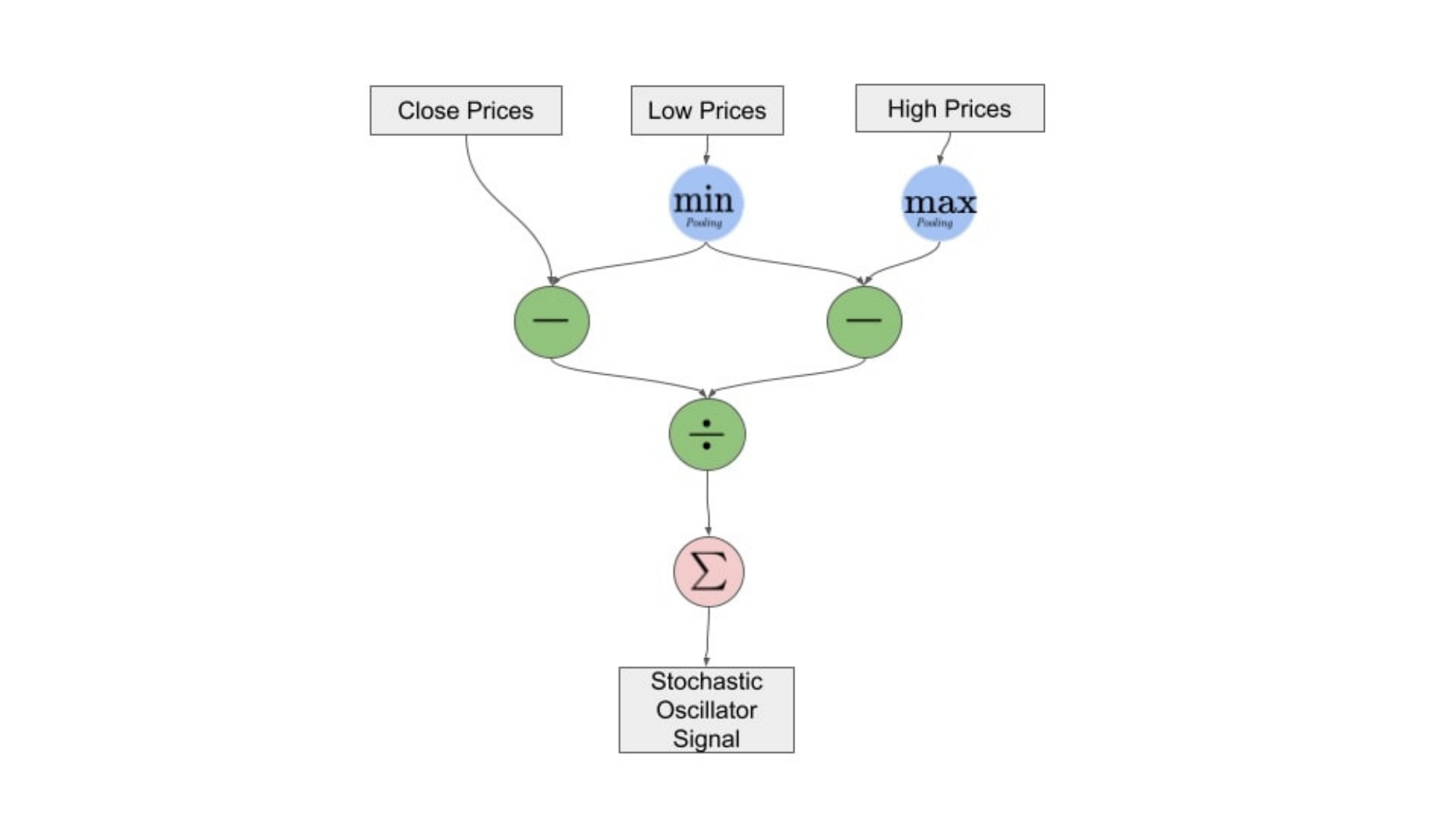}
\caption{Topology of Indicator Network Expressing the Stochastic Oscillator.}
\label{fig:stochTopology}
\end{figure}

The Stochastic Oscillator, which quantifies the relative position of the current closing price within a recent high-low range, is reformulated in the TINs framework as a sequence of layer operators computationally equivalent to its canonical definition. All operators are initialized directly from the classical formulation.

The architecture uses a maximum pooling operator to extract local highs and a complementary minimum pooling operator to identify local lows across a sliding window. These extrema are passed to a subtraction operator to compute the high-low spread, then to a division operator normalizing the current price position. This design preserves the momentum-based semantics of the original indicator while enabling adaptive parameter optimization.

\section{Commodity Channel Index (CCI) Indicator Network}

\begin{figure}[!t]
\centering
\includegraphics[width=0.98\linewidth]{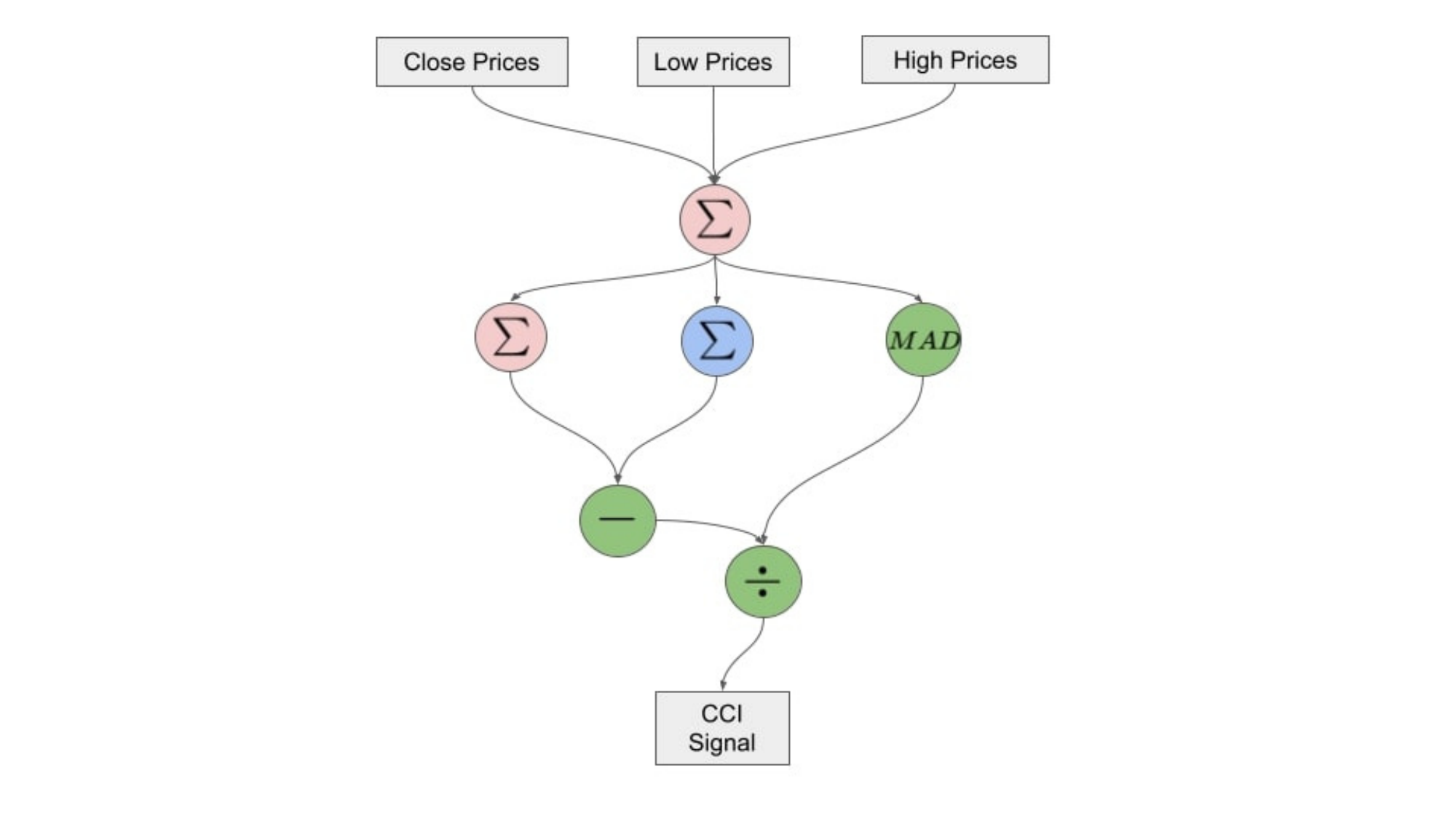}
\caption{Topology of Indicator Network Expressing the Commodity Channel Index (CCI).}
\label{fig:cci}
\end{figure}

The Commodity Channel Index (CCI), which measures the deviation of price from its dynamic mean, is implemented in the TINs framework as a multi-stage neural topology composed of differentiable layer operators. Each step—averaging, absolute deviation computation, and normalization—is initialized from the canonical CCI formula in accordance with the \textit{Layer Operator Equivalence} principle.

The architecture preserves the interpretability and cyclical trend detection of the original CCI, while enabling trainable adaptation and integration into larger neural trading systems.

\section{Risk-Adjusted Performance of TIN-MACD Across US30 Constituents}
\label{sec:tin_us30_performance}

This table reports the performance of three MACD-based strategies evaluated across the 30 constituents of the Dow Jones Industrial Average (US30): (i) the baseline rule-based MACD; (ii) a TIN implementation of MACD using price-only inputs; and (iii) a TIN-MACD variant incorporating both price and On-Balance Volume (OBV). 

All strategies are trained and tested on identical 52-day rolling windows of historical adjusted closing prices. The TIN architectures employ a single hidden layer of 26 units, initialized directly from the canonical MACD parameters $(12,26,9)$ via mathematically grounded layer operators. Model outputs are discretized into buy or sell signals for position-level simulation. 

Performance metrics are computed from daily portfolio returns, with the risk-free rate fixed at $R_f = 0$ for comparability. The Sharpe ratio is:
\begingroup
\setlength{\abovedisplayskip}{3pt}\setlength{\belowdisplayskip}{3pt}
\begin{equation}
\mathrm{Sharpe} = \frac{\mathbb{E}[R_p - R_f]}{\sigma_p}.
\end{equation}
\endgroup
Here, $\sigma_p$ denotes the standard deviation of daily portfolio returns.

The Sortino ratio replaces $\sigma_p$ with $\sigma_d$ (the standard deviation of downside returns):
\begingroup
\setlength{\abovedisplayskip}{3pt}\setlength{\belowdisplayskip}{3pt}
\begin{equation}
\mathrm{Sortino} = \frac{\mathbb{E}[R_p - R_f]}{\sigma_d}.
\end{equation}
\endgroup
These results show that topology-preserving, mathematically initialized TIN\text{-}MACD networks deliver improved risk-adjusted performance while retaining the interpretability of their classical counterparts.

\begin{longtable}{lcccccc}
\caption{Sharpe and Sortino Ratios for US30 Components under Different Strategies. \label{tab:ratios}}\\
\toprule
\textbf{Stock} &
\multicolumn{2}{c}{MACD} &
\multicolumn{2}{c}{TIN-Price} &
\multicolumn{2}{c}{TIN-Price+OBV} \\
\cmidrule(lr){2-3}\cmidrule(lr){4-5}\cmidrule(lr){6-7}
 & Sharpe & Sortino & Sharpe & Sortino & Sharpe & Sortino \\
\midrule
\endfirsthead

\toprule
\textbf{Stock} &
\multicolumn{2}{c}{MACD} &
\multicolumn{2}{c}{TIN-Price} &
\multicolumn{2}{c}{TIN-Price+OBV} \\
\cmidrule(lr){2-3}\cmidrule(lr){4-5}\cmidrule(lr){6-7}
 & Sharpe & Sortino & Sharpe & Sortino & Sharpe & Sortino \\
\midrule
\endhead

\midrule
\multicolumn{7}{r}{\textit{Continued on next page}} \\
\midrule
\endfoot

\bottomrule
\multicolumn{7}{l}{\scriptsize \textit{Paired $t$-test results (n=31) against MACD:}} \\
\multicolumn{7}{l}{\scriptsize Sharpe: \; TIN-Price \; $p=0.2113$, mean diff: $+0.1574$} \\
\multicolumn{7}{l}{\scriptsize Sharpe: \; TIN-Price+OBV \; $p=0.0202$, mean diff: $+0.3575$} \\
\multicolumn{7}{l}{\scriptsize Sortino: \; TIN-Price \; $p=0.0573$, mean diff: $+1.0371$} \\
\multicolumn{7}{l}{\scriptsize Sortino: \; TIN-Price+OBV \; $p=0.0485$, mean diff: $+1.0910$} \\
\endlastfoot

DJI & 0.5169 & 0.1110 & 0.8924 & \textbf{0.5898} & \textbf{0.9157} & 0.4025 \\
AAPL & -0.2256 & -0.0582 & 0.8429 & \textbf{9.3534} & \textbf{0.8648} & 0.1129 \\
AMGN & 0.3798 & 0.2039 & 0.0531 & 0.0161 & \textbf{1.3285} & \textbf{0.3677} \\
AXP & 0.6690 & 0.1445 & 1.0224 & \textbf{0.3792} & \textbf{1.1915} & 0.2382 \\
BA & 0.4686 & \textbf{0.2474} & 0.2530 & 0.1082 & \textbf{0.6674} & 0.1890 \\
CAT & \textbf{1.5083} & \textbf{0.8843} & 0.5231 & 0.1877 & 0.0823 & 0.0237 \\
CRM & 0.6798 & 0.2000 & \textbf{0.8578} & 0.2749 & 0.5355 & \textbf{0.3008} \\
CSCO & -0.6242 & -0.1807 & 0.5653 & 0.2873 & \textbf{0.8362} & \textbf{0.6769} \\
CVX & -0.4408 & -0.1330 & \textbf{0.0888} & \textbf{0.0244} & -0.6318 & -0.1848 \\
DIS & 0.3933 & 0.1639 & \textbf{1.1322} & \textbf{1.1893} & 0.6618 & 0.1961 \\
GS & 0.9239 & 0.2488 & \textbf{1.5282} & 0.8808 & 1.4734 & \textbf{10.0000} \\
HD & \textbf{1.3230} & \textbf{1.3142} & 1.2688 & 0.3211 & 0.6770 & 0.4600 \\
HON & -0.2040 & -0.0658 & \textbf{1.2913} & 0.4450 & 0.8676 & \textbf{10.0000} \\
IBM & 1.4143 & 0.4589 & \textbf{1.4441} & \textbf{10.0000} & 1.1917 & 0.7432 \\
INTC & 0.0797 & 0.0234 & -0.4112 & -0.1071 & \textbf{0.5725} & \textbf{0.0892} \\
JNJ & \textbf{1.2751} & \textbf{0.5938} & 0.6297 & 0.3778 & 0.0865 & 0.0140 \\
JPM & 1.4721 & 0.3364 & 1.0013 & 0.8129 & \textbf{1.7383} & \textbf{10.0000} \\
KO & 0.2361 & 0.1749 & 1.0249 & 1.1777 & \textbf{1.3536} & \textbf{1.7173} \\
MCD & -0.7480 & -0.3222 & \textbf{0.0702} & \textbf{0.0342} & 0.0597 & 0.0187 \\
MMM & \textbf{1.4331} & \textbf{1.5108} & 0.6809 & 0.6907 & 1.3890 & 1.3760 \\
MRK & -0.4265 & -0.1564 & \textbf{1.0123} & \textbf{10.0000} & 0.9221 & 0.6775 \\
MSFT & 0.4589 & 0.2643 & 0.6319 & \textbf{0.5277} & \textbf{0.9975} & 0.1789 \\
NKE & 0.4856 & 0.2316 & 0.1977 & 0.0324 & \textbf{0.6337} & \textbf{0.5177} \\
PG & -0.5370 & -0.1913 & -1.1695 & -0.2137 & \textbf{1.5681} & \textbf{0.8156} \\
TRV & \textbf{1.1391} & 0.5102 & 0.5179 & 0.2777 & 1.1310 & \textbf{0.5980} \\
UNH & \textbf{-0.3379} & \textbf{-0.0735} & -0.7096 & -0.1905 & -0.4814 & -0.0935 \\
V & 0.3161 & 0.1240 & 1.1878 & 0.7752 & \textbf{1.2160} & \textbf{1.0835} \\
VZ & -0.0227 & -0.0098 & 0.2355 & 0.1428 & \textbf{1.8223} & \textbf{0.5603} \\
WBA & -0.9361 & -0.3095 & \textbf{-0.4385} & \textbf{-0.0431} & -1.0199 & -0.1891 \\
WMT & \textbf{1.5726} & \textbf{0.9123} & 0.8208 & 0.7899 & 0.8936 & 0.2916 \\
RTX & 0.9161 & 0.4960 & \textbf{0.9936} & \textbf{0.6640} & 0.6967 & 0.2934 \\
\end{longtable}

Table~\ref{tab:ratios} summarizes the Sharpe and Sortino ratios for three MACD-based strategies evaluated across the 30 constituents of the Dow Jones Industrial Average (US30). The index itself is also tested as a separate trading target, denoted as DJI in the table. The paired t-test results provide formal statistical evidence regarding the relative performance of the proposed Technical Indicator Networks (TINs) compared with the canonical MACD. On Sharpe ratios, the TIN-Price variant yields a positive mean difference of +0.1574, though this improvement is not statistically significant (p = 0.2113). By contrast, the TIN-Price+OBV variant achieves a substantially larger mean improvement of +0.3575, which is statistically significant at the 5\% level (p = 0.0202). This indicates that augmenting the MACD operator with OBV within the TIN framework produces reliably higher risk-adjusted returns relative to the original MACD.

On Sortino ratios, the improvements are more pronounced. The TIN-Price variant produces an average increase of +1.0371 with borderline statistical significance (p = 0.0573), while the TIN-Price+OBV variant achieves a mean increase of +1.0910, which is statistically significant (p = 0.0485). In cases where no downside deviation is observed, the Sortino ratio is set to a default value of 10.0, reflecting a practically “perfect” risk profile with no underperformance relative to the target return. These results suggest that TINs not only improve average risk-adjusted returns but also reduce downside risk exposure, with the OBV-enhanced architecture providing the most consistent and statistically supported gains.

Taken together, these results suggest that Sharpe ratio enhancements from TINs, particularly the volume-augmented variant, are both economically meaningful and statistically supported. Sortino ratio improvements, while positive on average, are less consistent across the 30 constituents and therefore more fragile as evidence. Overall, the findings confirm that TINs provide systematic improvements over the canonical MACD, demonstrating that embedding traditional indicator logic within a topology-preserving neural design does not merely replicate existing heuristics but enhances their effectiveness in a risk-adjusted context.

It is noteworthy that while Sharpe ratio improvements of TINs over MACD are statistically significant, Sortino ratio enhancements, though significant in the OBV-augmented variant and borderline in the price-only configuration, remain less consistent across the US30 sample. Rather than a limitation, this observation highlights an important direction for future work: the current TIN formulation is primarily designed to preserve and enhance the risk-return balance as captured by Sharpe, but not explicitly optimized for downside risk, which is the focus of the Sortino metric. Future research therefore is possible to extends the TIN framework by incorporating loss-sensitive reinforcement learning objectives or downside-risk-aware operator designs, which is possible to further improve Sortino performance and enhance robustness under adverse market conditions. Future extensions draw from recent ML in finance to enhance robustness under regime shifts~\cite{Nazareth2023ESWA_Review,Yun2025ESWA_TechIndicators}, and to connect downside-risk-aware objectives with the TIN operator design.


\section*{CRediT authorship contribution statement}
The author conducted this research independently, without external funding or institutional support, and confirms being the sole contributor to all aspects of the work, including conceptualization, methodology, software, analysis, visualization, and writing. The author has read and approved the final version of the manuscript and accepts responsibility for its accuracy and integrity.

\section*{Declaration of competing interest}
The author has no other financial or personal competing interests to disclose.

\section*{Data availability}
The data supporting the findings of this study are publicly available from the Yahoo Finance API via the \texttt{yfinance} library. The processed datasets generated during the current study are available on reasonable request.

\section*{Declaration of Generative AI and AI-assisted technologies in the writing process}
During the preparation of this work the author used ChatGPT (OpenAI) for language editing.
After using this tool, the author reviewed and edited the content as needed and takes full responsibility for the content of the publication.

\section*{Funding}
This research did not receive any specific grant from funding agencies in the public, commercial, or not-for-profit sectors, and was conducted independently by the author.

\bibliography{MyLibrary}

\begin{thebibliography}{25}
\expandafter\ifx\csname natexlab\endcsname\relax\def\natexlab#1{#1}\fi
\providecommand{\url}[1]{\texttt{#1}}
\providecommand{\href}[2]{#2}
\providecommand{\path}[1]{#1}
\providecommand{\DOIprefix}{doi:}
\providecommand{\ArXivprefix}{arXiv:}
\providecommand{\URLprefix}{URL: }
\providecommand{\Pubmedprefix}{pmid:}
\providecommand{\doi}[1]{\href{http://dx.doi.org/#1}{\path{#1}}}
\providecommand{\Pubmed}[1]{\href{pmid:#1}{\path{#1}}}
\providecommand{\bibinfo}[2]{#2}
\ifx\xfnm\relax \def\xfnm[#1]{\unskip,\space#1}\fi
\bibitem[{Al-Yahya et~al.(2021)Al-Yahya, Mehmood, Albeshri, Damiani \& Song}]{alyahya2021}
\bibinfo{author}{Al-Yahya, M.}, \bibinfo{author}{Mehmood, R.}, \bibinfo{author}{Albeshri, A.}, \bibinfo{author}{Damiani, E.}, \& \bibinfo{author}{Song, H.} (\bibinfo{year}{2021}).
\newblock \bibinfo{title}{A deep learning approach for forecasting cryptocurrencies using arima-lstm}.
\newblock {\it \bibinfo{journal}{IEEE Access}\/},  {\it \bibinfo{volume}{9}\/}, \bibinfo{pages}{97324--97338}. \DOIprefix\doi{10.1109/ACCESS.2021.3095101}.
\bibitem[{Bao et~al.(2017)Bao, Yue \& Rao}]{bao2017}
\bibinfo{author}{Bao, W.}, \bibinfo{author}{Yue, J.}, \& \bibinfo{author}{Rao, Y.} (\bibinfo{year}{2017}).
\newblock \bibinfo{title}{A deep learning framework for financial time series using stacked autoencoders and long short-term memory}.
\newblock {\it \bibinfo{journal}{PLoS ONE}\/},  {\it \bibinfo{volume}{12}\/}, \bibinfo{pages}{e0180944}. \DOIprefix\doi{10.1371/journal.pone.0180944}.
\bibitem[{Chudziak \& {others}(2023)}]{Chudziak2023ESWA}
\bibinfo{author}{Chudziak, A.}, \& \bibinfo{author}{{others}} (\bibinfo{year}{2023}).
\newblock \bibinfo{title}{Predictability of stock returns using neural networks}.
\newblock {\it \bibinfo{journal}{Expert Systems with Applications}\/},  {\it \bibinfo{volume}{216}\/}, \bibinfo{pages}{119409}. \URLprefix \url{https://www.sciencedirect.com/science/article/abs/pii/S0957417422022217}.
\bibitem[{Deng et~al.(2017)Deng, Bao, Kong, Ren \& Dai}]{Deng2017TNNLS_DeepDirectRL}
\bibinfo{author}{Deng, Y.}, \bibinfo{author}{Bao, F.}, \bibinfo{author}{Kong, Y.}, \bibinfo{author}{Ren, Z.}, \& \bibinfo{author}{Dai, Q.} (\bibinfo{year}{2017}).
\newblock \bibinfo{title}{Deep direct reinforcement learning for financial signal representation and trading}.
\newblock {\it \bibinfo{journal}{IEEE Transactions on Neural Networks and Learning Systems}\/},  {\it \bibinfo{volume}{28}\/}, \bibinfo{pages}{653--664}. \URLprefix \url{https://doi.org/10.1109/TNNLS.2016.2522401}.
\bibitem[{Ding et~al.(2022)Ding, Pan, Cui \& Huang}]{ding2022}
\bibinfo{author}{Ding, Q.}, \bibinfo{author}{Pan, L.}, \bibinfo{author}{Cui, S.}, \& \bibinfo{author}{Huang, Q.} (\bibinfo{year}{2022}).
\newblock \bibinfo{title}{Drl-based portfolio management with transaction cost}.
\newblock {\it \bibinfo{journal}{Expert Systems with Applications}\/},  {\it \bibinfo{volume}{208}\/}, \bibinfo{pages}{118170}. \DOIprefix\doi{10.1016/j.eswa.2022.118170}.
\bibitem[{Felizardo \& {others}(2022)}]{Felizardo2022ESWA_RL}
\bibinfo{author}{Felizardo, L.~K.}, \& \bibinfo{author}{{others}} (\bibinfo{year}{2022}).
\newblock \bibinfo{title}{Outperforming algorithmic trading reinforcement learning baselines}.
\newblock {\it \bibinfo{journal}{Expert Systems with Applications}\/},  {\it \bibinfo{volume}{200}\/}, \bibinfo{pages}{117001}. \URLprefix \url{https://www.sciencedirect.com/science/article/abs/pii/S0957417422006339}.
\bibitem[{Fischer \& Krauss(2018)}]{fischer2018}
\bibinfo{author}{Fischer, T.}, \& \bibinfo{author}{Krauss, C.} (\bibinfo{year}{2018}).
\newblock \bibinfo{title}{Deep learning with long short-term memory networks for financial market predictions}.
\newblock {\it \bibinfo{journal}{European Journal of Operational Research}\/},  {\it \bibinfo{volume}{270}\/}, \bibinfo{pages}{654--669}. \DOIprefix\doi{10.1016/j.ejor.2018.04.016}.
\bibitem[{Gajamannage \& {others}(2023)}]{Gajamannage2023ESWA}
\bibinfo{author}{Gajamannage, K.}, \& \bibinfo{author}{{others}} (\bibinfo{year}{2023}).
\newblock \bibinfo{title}{Real-time forecasting of time series in financial markets}.
\newblock {\it \bibinfo{journal}{Expert Systems with Applications}\/},  {\it \bibinfo{volume}{213}\/}, \bibinfo{pages}{118834}. \URLprefix \url{https://www.sciencedirect.com/science/article/abs/pii/S0957417423003809}.
\bibitem[{G{\"u}lmez(2023)}]{Gulmez2023ESWA}
\bibinfo{author}{G{\"u}lmez, B.} (\bibinfo{year}{2023}).
\newblock \bibinfo{title}{Stock price prediction with optimized deep lstm network}.
\newblock {\it \bibinfo{journal}{Expert Systems with Applications}\/},  {\it \bibinfo{volume}{228}\/}, \bibinfo{pages}{120129}. \URLprefix \url{https://www.sciencedirect.com/science/article/pii/S0957417423008485}.
\bibitem[{Kang(2021)}]{kang_improving_2021}
\bibinfo{author}{Kang, B.-K.} (\bibinfo{year}{2021}).
\newblock \bibinfo{title}{Improving {MACD} {Technical} {Analysis} by {Optimizing} {Parameters} and {Modifying} {Trading} {Rules}: {Evidence} from the {Japanese} {Nikkei} 225 {Futures} {Market}}.
\newblock {\it \bibinfo{journal}{Journal of Risk and Financial Management}\/},  {\it \bibinfo{volume}{14}\/}, \bibinfo{pages}{37}. \URLprefix \url{https://www.mdpi.com/1911-8074/14/1/37}. \DOIprefix\doi{10.3390/jrfm14010037}.
\bibitem[{Kehinde \& {others}(2023)}]{Kehinde2023ESWA_Scientometric}
\bibinfo{author}{Kehinde, T.~O.}, \& \bibinfo{author}{{others}} (\bibinfo{year}{2023}).
\newblock \bibinfo{title}{Scientometric review and analysis of recent approaches to financial forecasting}.
\newblock {\it \bibinfo{journal}{Expert Systems with Applications}\/},  {\it \bibinfo{volume}{225}\/}, \bibinfo{pages}{120168}. \URLprefix \url{https://www.sciencedirect.com/science/article/abs/pii/S095741742202317X}.
\bibitem[{Kwon et~al.(2022)Kwon, Son \& Kim}]{kwon2022}
\bibinfo{author}{Kwon, G.}, \bibinfo{author}{Son, A.~R.}, \& \bibinfo{author}{Kim, D.~P.} (\bibinfo{year}{2022}).
\newblock \bibinfo{title}{A robust time-series validation approach for financial forecasting with deep neural networks}.
\newblock {\it \bibinfo{journal}{Applied Soft Computing}\/},  {\it \bibinfo{volume}{116}\/}, \bibinfo{pages}{108197}. \DOIprefix\doi{10.1016/j.asoc.2021.108197}.
\bibitem[{Liu \& {others}(2024)}]{Liu2024ESWA_NEV}
\bibinfo{author}{Liu, X.}, \& \bibinfo{author}{{others}} (\bibinfo{year}{2024}).
\newblock \bibinfo{title}{Stock price prediction for new energy vehicle companies using deep learning approaches}.
\newblock {\it \bibinfo{journal}{Expert Systems with Applications}\/},  {\it \bibinfo{volume}{237}\/}, \bibinfo{pages}{121616}. \URLprefix \url{https://www.sciencedirect.com/science/article/abs/pii/S0957417424016543}.
\bibitem[{Majidi \& {others}(2024)}]{Majidi2024ESWA_ContinuousRL}
\bibinfo{author}{Majidi, N.}, \& \bibinfo{author}{{others}} (\bibinfo{year}{2024}).
\newblock \bibinfo{title}{Algorithmic trading using continuous action space deep reinforcement learning}.
\newblock {\it \bibinfo{journal}{Expert Systems with Applications}\/},  {\it \bibinfo{volume}{235}\/}, \bibinfo{pages}{121245}. \URLprefix \url{https://www.sciencedirect.com/science/article/abs/pii/S0957417423017475}.
\bibitem[{Mnih et~al.(2016)Mnih, Badia, Mirza, Graves, Lillicrap, Harley, Silver \& Kavukcuoglu}]{Mnih2016A3C}
\bibinfo{author}{Mnih, V.}, \bibinfo{author}{Badia, A.~P.}, \bibinfo{author}{Mirza, M.}, \bibinfo{author}{Graves, A.}, \bibinfo{author}{Lillicrap, T.}, \bibinfo{author}{Harley, T.}, \bibinfo{author}{Silver, D.}, \& \bibinfo{author}{Kavukcuoglu, K.} (\bibinfo{year}{2016}).
\newblock \bibinfo{title}{Asynchronous methods for deep reinforcement learning}.
\newblock In {\it \bibinfo{booktitle}{Proceedings of the 33rd International Conference on Machine Learning (ICML)}\/} (pp. \bibinfo{pages}{1928--1937}).
\newblock \URLprefix \url{https://arxiv.org/abs/1602.01783}.
\bibitem[{Mnih et~al.(2015)Mnih, Kavukcuoglu, Silver, Rusu, Veness, Bellemare, Graves, Riedmiller, Fidjeland, Ostrovski, Petersen, Beattie, Sadik, Antonoglou, King, Kumaran, Wierstra, Legg \& Hassabis}]{Mnih2015DQN}
\bibinfo{author}{Mnih, V.}, \bibinfo{author}{Kavukcuoglu, K.}, \bibinfo{author}{Silver, D.}, \bibinfo{author}{Rusu, A.~A.}, \bibinfo{author}{Veness, J.}, \bibinfo{author}{Bellemare, M.~G.}, \bibinfo{author}{Graves, A.}, \bibinfo{author}{Riedmiller, M.}, \bibinfo{author}{Fidjeland, A.~K.}, \bibinfo{author}{Ostrovski, G.}, \bibinfo{author}{Petersen, S.}, \bibinfo{author}{Beattie, C.}, \bibinfo{author}{Sadik, A.}, \bibinfo{author}{Antonoglou, I.}, \bibinfo{author}{King, H.}, \bibinfo{author}{Kumaran, D.}, \bibinfo{author}{Wierstra, D.}, \bibinfo{author}{Legg, S.}, \& \bibinfo{author}{Hassabis, D.} (\bibinfo{year}{2015}).
\newblock \bibinfo{title}{Human-level control through deep reinforcement learning}.
\newblock {\it \bibinfo{journal}{Nature}\/},  {\it \bibinfo{volume}{518}\/}, \bibinfo{pages}{529--533}. \URLprefix \url{https://doi.org/10.1038/nature14236}. \DOIprefix\doi{10.1038/nature14236}.
\bibitem[{Mostafavi \& {others}(2025)}]{Yun2025ESWA_TechIndicators}
\bibinfo{author}{Mostafavi, S.~M.}, \& \bibinfo{author}{{others}} (\bibinfo{year}{2025}).
\newblock \bibinfo{title}{Key technical indicators for stock market prediction}.
\newblock {\it \bibinfo{journal}{Expert Systems with Applications}\/},  {\it \bibinfo{volume}{259}\/}, \bibinfo{pages}{125047}. \URLprefix \url{https://www.sciencedirect.com/science/article/pii/S2666827025000143}.
\bibitem[{Nabipour et~al.(2020)Nabipour, Nayyeri, Jabani, Mosavi \& Salwana}]{nabipour2020}
\bibinfo{author}{Nabipour, M.}, \bibinfo{author}{Nayyeri, P.}, \bibinfo{author}{Jabani, H.}, \bibinfo{author}{Mosavi, A.}, \& \bibinfo{author}{Salwana, E.} (\bibinfo{year}{2020}).
\newblock \bibinfo{title}{Predicting stock market trends using machine learning and deep learning: The case of tehran stock exchange}.
\newblock {\it \bibinfo{journal}{IEEE Access}\/},  {\it \bibinfo{volume}{8}\/}, \bibinfo{pages}{150199--150212}. \DOIprefix\doi{10.1109/ACCESS.2020.3015966}.
\bibitem[{Nazareth \& {others}(2023)}]{Nazareth2023ESWA_Review}
\bibinfo{author}{Nazareth, N.}, \& \bibinfo{author}{{others}} (\bibinfo{year}{2023}).
\newblock \bibinfo{title}{Financial applications of machine learning: A literature review}.
\newblock {\it \bibinfo{journal}{Expert Systems with Applications}\/},  {\it \bibinfo{volume}{215}\/}, \bibinfo{pages}{119113}. \URLprefix \url{https://www.sciencedirect.com/science/article/abs/pii/S0957417423001410}.
\bibitem[{Ni et~al.(2021)Ni, Yin \& Wang}]{ni2021}
\bibinfo{author}{Ni, J.}, \bibinfo{author}{Yin, S.}, \& \bibinfo{author}{Wang, F.} (\bibinfo{year}{2021}).
\newblock \bibinfo{title}{An ensemble deep learning model for stock price prediction}.
\newblock {\it \bibinfo{journal}{Neural Computing and Applications}\/},  {\it \bibinfo{volume}{33}\/}, \bibinfo{pages}{8397--8407}. \DOIprefix\doi{10.1007/s00521-020-05656-8}.
\bibitem[{Ricchiuti \& {others}(2025)}]{Ricchiuti2025ESWA_AdvisorNN}
\bibinfo{author}{Ricchiuti, F.}, \& \bibinfo{author}{{others}} (\bibinfo{year}{2025}).
\newblock \bibinfo{title}{An advisor neural network framework using lstm-based informative stock analysis}.
\newblock {\it \bibinfo{journal}{Expert Systems with Applications}\/},  {\it \bibinfo{volume}{260}\/}, \bibinfo{pages}{125173}. \URLprefix \url{https://www.sciencedirect.com/science/article/pii/S0957417424021663}.
\bibitem[{Schulman et~al.(2016)Schulman, Moritz, Levine, Jordan \& Abbeel}]{Schulman2015GAE}
\bibinfo{author}{Schulman, J.}, \bibinfo{author}{Moritz, P.}, \bibinfo{author}{Levine, S.}, \bibinfo{author}{Jordan, M.}, \& \bibinfo{author}{Abbeel, P.} (\bibinfo{year}{2016}).
\newblock \bibinfo{title}{High-dimensional continuous control using generalized advantage estimation}.
\newblock In {\it \bibinfo{booktitle}{International Conference on Learning Representations (ICLR)}\/}.
\newblock \URLprefix \url{https://arxiv.org/abs/1506.02438}.
\bibitem[{Th{\'e}ate \& Ernst(2021)}]{Theate2021ESWA_DRL}
\bibinfo{author}{Th{\'e}ate, T.}, \& \bibinfo{author}{Ernst, D.} (\bibinfo{year}{2021}).
\newblock \bibinfo{title}{An application of deep reinforcement learning to algorithmic trading}.
\newblock {\it \bibinfo{journal}{Expert Systems with Applications}\/},  {\it \bibinfo{volume}{173}\/}, \bibinfo{pages}{114632}. \URLprefix \url{https://www.sciencedirect.com/science/article/abs/pii/S0957417421000737}.
\bibitem[{Williams(1992)}]{Williams1992REINFORCE}
\bibinfo{author}{Williams, R.~J.} (\bibinfo{year}{1992}).
\newblock \bibinfo{title}{Simple statistical gradient-following algorithms for connectionist reinforcement learning}.
\newblock {\it \bibinfo{journal}{Machine Learning}\/},  {\it \bibinfo{volume}{8}\/}, \bibinfo{pages}{229--256}. \URLprefix \url{https://doi.org/10.1007/BF00992696}. \DOIprefix\doi{10.1007/BF00992696}.
\bibitem[{Yun \& {others}(2023)}]{Yun2023ESWA_GA}
\bibinfo{author}{Yun, K.-K.}, \& \bibinfo{author}{{others}} (\bibinfo{year}{2023}).
\newblock \bibinfo{title}{Interpretable stock price forecasting model using genetic algorithms and machine learning}.
\newblock {\it \bibinfo{journal}{Expert Systems with Applications}\/},  {\it \bibinfo{volume}{226}\/}, \bibinfo{pages}{120133}. \URLprefix \url{https://www.sciencedirect.com/science/article/abs/pii/S0957417422018218}.

\end{thebibliography}

\end{document}